\documentclass[letterpaper]{article} % DO NOT CHANGE THIS
\usepackage[preprint]{aaai2027} % preprint: de-anonymized, drops the review-only copyright slug (arXiv build)
\usepackage[hyphens]{url} % DO NOT CHANGE THIS
\usepackage{graphicx} % DO NOT CHANGE THIS
\usepackage{natbib} % DO NOT CHANGE THIS AND DO NOT ADD ANY OPTIONS TO IT
\usepackage{caption} % DO NOT CHANGE THIS AND DO NOT ADD ANY OPTIONS TO IT
\usepackage{amsmath}
\usepackage{amssymb}
\usepackage{booktabs}
\usepackage{multirow}
\usepackage{makecell}
\usepackage{subcaption}

\usepackage{tcolorbox}

\definecolor{linkblue}{RGB}{40,96,214}
\newcommand{\reflinkicon}[1]{\raisebox{-0.22em}{\includegraphics[height=1.05em]{#1}}}

\usepackage{newfloat}
\DeclareFloatingEnvironment[
 fileext=lop,
 listname={List of Prompts},
 name=Prompt,
 placement=tbp
]{prompt}

\usepackage[hidelinks,breaklinks=true]{hyperref}

\title{VC-Tooler: Learning Compositional and Adaptive Visual Tool Use}

\author{
    Yizheng Wu,\quad
    Jiashen Hua\corresponding,\quad
    Bing Deng,\quad
    Jieping Ye
}
\affiliations{
    Alibaba Group\\[6pt]
    \normalsize
    \reflinkicon{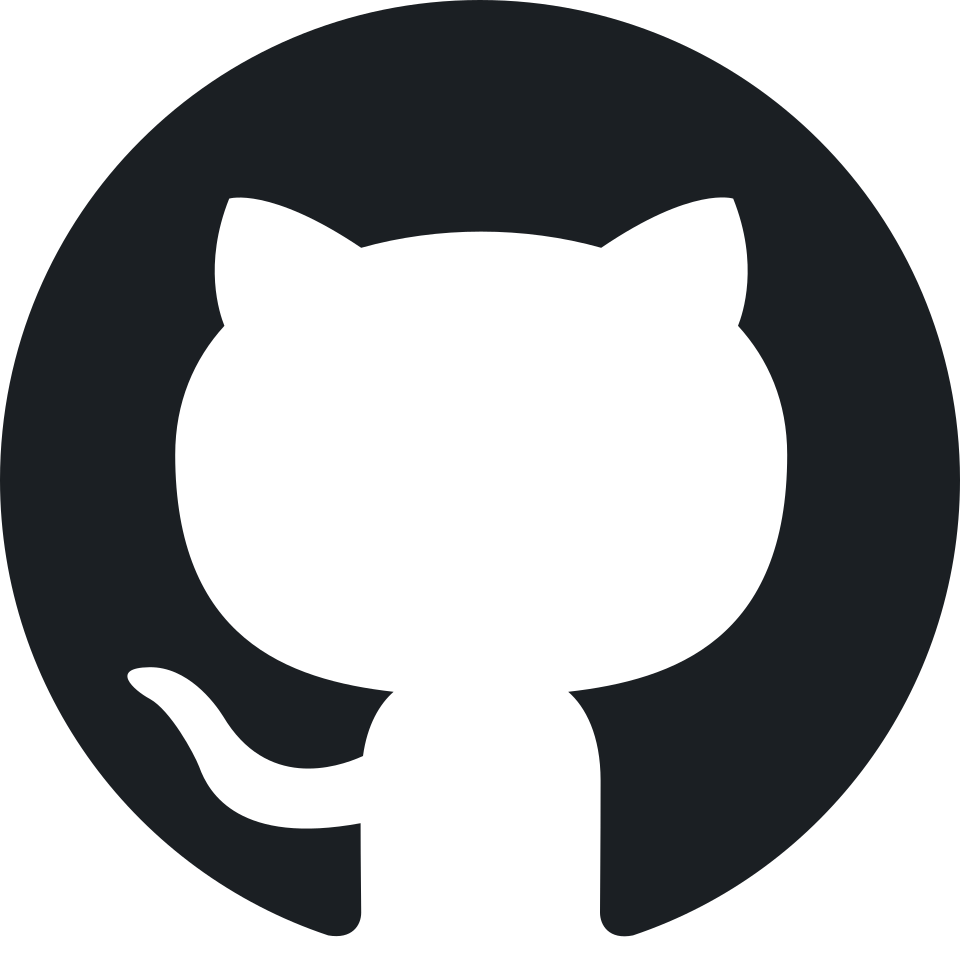}\hspace{0.4em}\textcolor{linkblue}{\href{https://w1zheng.github.io/VC-Tooler}{\nolinkurl{https://w1zheng.github.io/VC-Tooler}}}\\[6pt]
    \reflinkicon{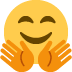}\hspace{0.35em}\href{https://huggingface.co/datasets/5551z/VC-Tooler-SFT}{\textcolor{linkblue}{VC-Tooler-SFT}}\hspace{1.6em}%
    \reflinkicon{figure/icon-hf.png}\hspace{0.35em}\href{https://huggingface.co/datasets/5551z/VC-Tooler-RL}{\textcolor{linkblue}{VC-Tooler-RL}}\hspace{1.6em}%
    \reflinkicon{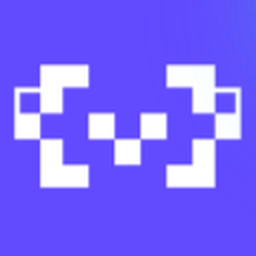}\hspace{0.35em}\href{https://modelscope.cn/datasets/W1zheng/VC-Tooler-SFT}{\textcolor{linkblue}{VC-Tooler-SFT}}\hspace{1.6em}%
    \reflinkicon{figure/icon-ms.png}\hspace{0.35em}\href{https://modelscope.cn/datasets/W1zheng/VC-Tooler-RL}{\textcolor{linkblue}{VC-Tooler-RL}}%
}

\begin{document}
\maketitle

\begin{abstract}
Agentic multimodal reasoning extends passive image understanding by allowing VLMs to actively acquire and refine visual evidence through visual tool interactions. Effective visual tool use requires three capabilities: grounding tool calls in visual context, composing tools across multiple steps, and adapting reasoning to tool-returned observations. However, existing approaches largely focus on grounding within fixed tool spaces and rigid invocation patterns, leaving composition and adaptation insufficiently addressed. We present VC-Tooler, which learns visual tool use as a compositional and adaptive capability. To this end, we first build a trajectory bank through a hierarchical synthesis pipeline covering three capability levels: single-tool grounding, multi-tool composition, and diverse tool contexts and interfaces. We then train the model in two stages: a supervised cold start that establishes these capabilities, followed by reinforcement learning that encourages accurate, efficient, and context-aware visual tool use. VC-Tooler achieves state-of-the-art performance among open-source models on both general-purpose and agentic benchmarks, including $95.8\%$ on V* and $35.3\%$ on VTC-Bench, and shows promising transfer under richer tool settings at inference time.
\end{abstract}

\section{Introduction}
\begin{figure}[t!]
 \centering
 \includegraphics[width=0.95\columnwidth]{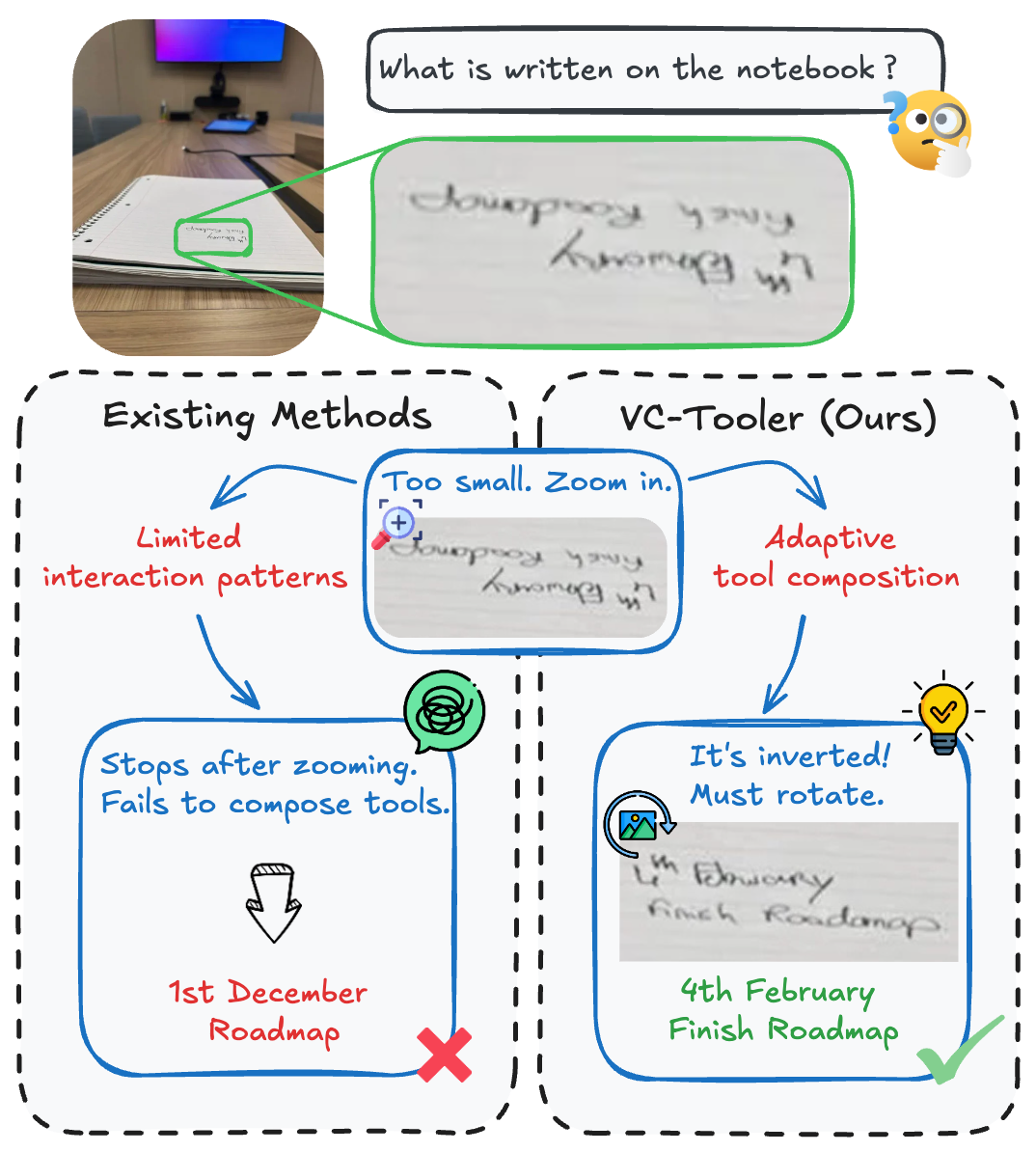}
 \caption{Existing methods often ground a single familiar tool call but fail to compose multiple tools or adapt to new observations. VC-Tooler continues the trajectory by composing tools and adapting subsequent actions to tool-returned observations.
}
 \label{fig:fig1}
 \vspace{-0.6cm}
\end{figure}

Recent progress in vision-language models (VLMs) has increasingly moved beyond passive image understanding toward agentic multimodal reasoning, where models actively use external tools to support multi-step problem solving~\cite{su2025thinking,wei2026agentic}. In this setting, visual tools are central: some transform visual evidence, such as zooming~\cite{wang2025pixel,zheng2025deepeyes} or rotating~\cite{zhang2025thyme,xu2025vacot} an image to reveal fine-grained details; others provide complementary signals, such as retrieval~\cite{hong2025deepeyesv2,geng2025webwatcher} or imagination~\cite{chern2025thinking,li2025zebra}, beyond the original input. By using such tools, VLMs can reason not only about an image, but also through the controlled acquisition and refinement of visual evidence.

Effective visual tool use demands more than access to tools. VLMs must not only ground individual tool calls in visual evidence and tool schemas, but also compose multiple tools across steps and adapt their subsequent decisions to tool-returned observations and changing tool interfaces. Despite rapid progress, however, most methods are trained and evaluated on a small, fixed set of tools and interaction patterns, and struggle to compose or adapt their tool use once the tool space changes~\cite{zhu2026vtc,li2025tir}, as illustrated in Fig.~\ref{fig:fig1}. This limitation largely traces back to a data bottleneck: learning visual tool use as a general capability calls for training data that spans diverse tools and multi-step interactions, yet such trajectories are especially hard to obtain. Whereas text agents can draw on abundant APIs and function-calling corpora~\cite{liu2024apigen,yang2025toolmind,feng2025retool}, visual agents need executable environments and trajectories grounded in image-state transitions. As a result, when supervision is confined to a handful of tools, models tend to memorize fixed invocation patterns instead of acquiring transferable tool-use skills.

We address this bottleneck with VC-Tooler, a multimodal model for compositional and adaptive visual tool use. Our key idea is to treat visual tool use as a capability to be learned, rather than optimizing only for the final answer under a fixed tool environment. To this end, we identify three core skills required for effective visual tool use: grounding, composition, and adaptation; and train them jointly through scalable trajectory synthesis and a two-stage learning framework. Specifically, we construct a hierarchical trajectory pipeline that provides supervision at three levels: single-tool grounding, multi-tool composition, and diverse tool contexts. This produces a large corpus spanning hundreds of visual tool interfaces and contextual variants, encouraging generalization beyond fixed schemas and invocation forms. We then train the model in two stages: a supervised cold start that establishes these capabilities, followed by reinforcement learning that encourages accurate and effective use of tool-returned observations during reasoning.

Experimental results show that VC-Tooler achieves state-of-the-art performance among open-source models on eight general-purpose and agentic benchmarks. In particular, it attains $95.8\%$ on V*~\cite{wu2024v}, demonstrating strong fine-grained visual perception, and $35.3\%$ on VTC-Bench~\cite{zhu2026vtc}, indicating substantial gains in compositional visual tool use. VC-Tooler also shows promising zero-shot transfer to richer tool settings with novel tool interfaces, suggesting generalization beyond fixed tool-specific invocation patterns. Together, these results suggest that visual tool use can be learned as a compositional and adaptive capability rather than a collection of fixed calling patterns.

\section{Related Work}
\noindent \textbf{Multimodal Reasoning and Chain-of-Thought.}
Recent vision-language models have advanced from perception-oriented tasks toward more complex multimodal reasoning, marking a transition from System 1 to System 2~\citep{kahneman2011thinking}. Inspired by chain-of-thought reasoning in language models, prior work extends step-by-step reasoning, rationale generation, self-consistency, reflection, and verification to multimodal settings~\citep{wei2025open,xu2025llava,yang2025r1,pan2026through}. Recent progress in reasoning-oriented language models, particularly those improved through reinforcement learning such as OpenAI o1~\citep{jaech2024openai} and DeepSeek-R1~\citep{guo2025deepseek}, further underscores the importance of long-horizon and deliberate reasoning for complex inference. However, these methods still reason primarily over fixed visual inputs, and thus remain limited to passive perception without actively acquiring or transforming visual evidence~\citep{su2025thinking}.

\noindent \textbf{Agentic Visual Reasoning and Visual Tool Use.}
To overcome the limits of static perception, recent work has shifted toward agentic visual reasoning, where models interact with external tools to inspect, manipulate, or augment visual evidence~\citep{su2025thinking}. Existing efforts have progressed from scripted tool orchestration~\citep{zhao2025pyvision,li2025insight}, to systems equipped with task-specific visual tools~\citep{wang2025pixel,hong2025deepeyesv2,zhang2025thyme}, and more recently to improving multi-turn tool-calling via supervised fine-tuning or reinforcement learning~\citep{wang2025adatooler,yang2025deep,song2026adareasoner,lu2026scaling}.

A central challenge in this setting is to learn transferable tool-use capability rather than behavior tied to a fixed tool set. In LLM-based agents, this challenge has been partly addressed through large-scale tool-augmented training, diverse API collections, and agentic RL~\citep{liu2024apigen,yang2025toolmind,feng2025retool,zeng2026glm}. In the visual domain, however, visual signals make tool definition and environment construction substantially harder, which has also encouraged the use of more flexible programmatic interfaces~\citep{zhang2025thyme,hong2025deepeyesv2,google2026agentic}. Consequently, existing visual tool-use methods are still largely restricted to grounding in familiar tool spaces and known interaction patterns, with limited robustness when faced with compositional and adaptive scenarios. Although recent benchmarks have begun to evaluate these capabilities~\citep{zhu2026vtc,li2025tir}, systematic training for compositional and adaptive visual tool use remains underexplored. This gap motivates the development of VC-Tooler.

\section{Preliminary: Visual Tool Use}
\label{sec:preliminary}

We study an agentic multimodal model equipped with visual tool-use capabilities that operates under the standard ReAct-style~\cite{yao2022react} paradigm, in which language reasoning is interleaved with tool use to solve a query. Formally, given a user query $q$, a visual input $v$, and a set of available tools $\mathcal{T}=\{t_1,\dots,t_n\}$, the model generates a $K$-turn trajectory $\tau = (s_1, a_1, o_1, \dots, s_K, a_K, o_K), $ where each step consists of an intermediate reasoning state $s_k$, an action $a_k=(t_k, u_k)$, and the resulting observation $o_k$. Each tool $t_i \in \mathcal{T}$ is described by a schema specifying its interface and expected arguments. At step $k$, $t_k \in \mathcal{T}$ denotes the selected tool, and $u_k$ denotes the arguments predicted according to the corresponding tool schema. Executing the action $a_k$ yields an observation $o_k$, which may include transformed visual evidence, retrieved content, or other auxiliary outputs. After multiple interactions, the model produces a final answer $y$.

Under this formulation, an agentic multimodal model reasons not only over a fixed input but also over an evolving visual workspace that is progressively updated by tool-returned observations. This requires three core visual tool-use capabilities:
(1)~\textbf{visual tool grounding}, interpreting tool schemas, grounding tool calls in the visual input, and predicting valid arguments to execute individual tools;
(2)~\textbf{compositional tool use}, chaining tool calls to progressively acquire and refine visual evidence; and
(3)~\textbf{adaptive tool use}, re-grounding subsequent decisions in observations returned by tools and generalizing across heterogeneous tool interfaces.
These capabilities define the design objectives of VC-Tooler and motivate our training framework.

\section{Methodology}
\label{sec:methodology}
\begin{figure*}[t!]
 \centering
 \includegraphics[width=0.98\textwidth]{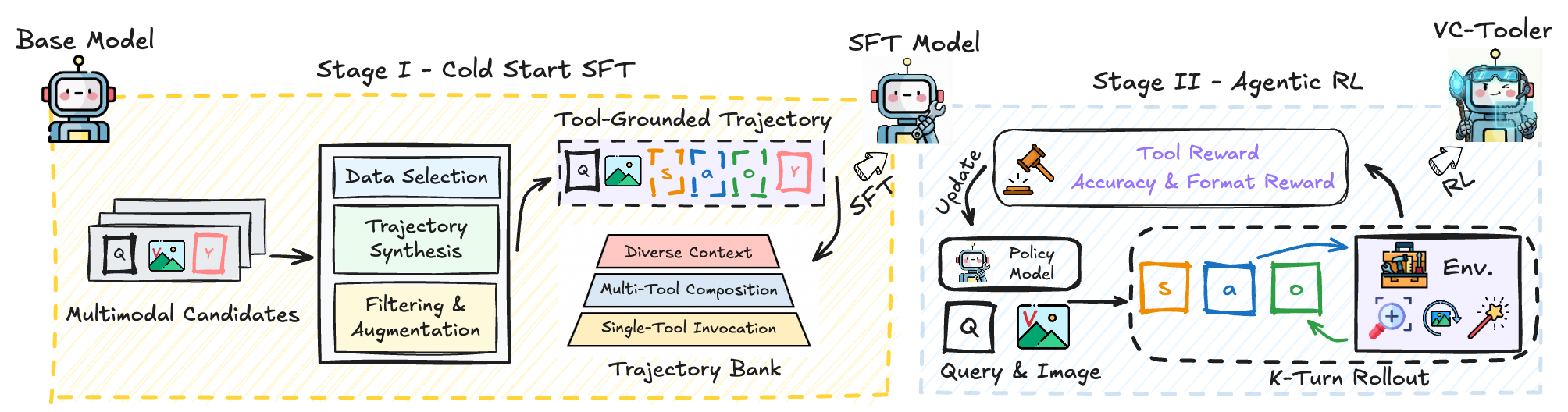}
 \caption{\textbf{Overview of VC-Tooler.} A two-stage framework with hierarchical trajectory supervision for cold start and tool-reward RL for policy refinement.}
 \label{fig:overview}
 \vspace{-0.2cm}
\end{figure*}

\subsection{Hierarchical Trajectory Synthesis}
\label{sec:synthesis}
\begin{figure*}[t!]
 \centering
 \includegraphics[width=0.9\textwidth]{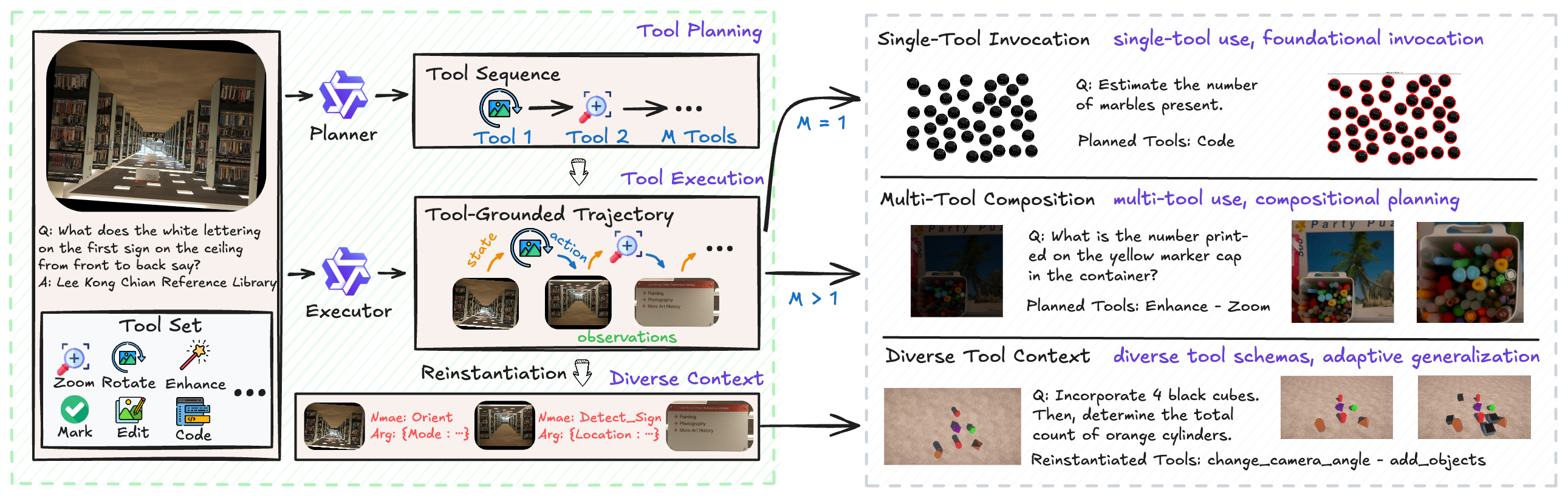}
 \caption{Trajectory synthesis via plan-then-execute and diverse tool-context reinstantiating.}
 \label{fig:trajectory}
  \vspace{-0.4cm}
\end{figure*}

Building on the three capability requirements identified in Sec.~\ref{sec:preliminary}, VC-Tooler comprises two components (Fig.~\ref{fig:overview}): (1) a hierarchical trajectory synthesis pipeline (Sec.~\ref{sec:synthesis}) that transforms large-scale raw VQA data into diverse tool-use trajectories at three levels, covering basic visual tool grounding while placing particular emphasis on compositional and adaptive tool use; and (2) a two-stage training procedure (Sec.~\ref{sec:training}) that first performs a supervised cold start on the synthesized trajectories and then further refines the resulting policy model through agentic reinforcement learning.

\subsubsection{Data Selection.}
We construct candidate samples for trajectory synthesis under three criteria: \emph{diversity}, by sampling from large-scale multimodal datasets such as LLaVA-OneVision~\cite{an2025llava,li2024llava}, DeepVision~\cite{sun2026deepvision}, and VisualProbe~\cite{lai2025mini}, covering varied visual scenes and task types; \emph{verifiability}, by retaining instances with reliably checkable answers; and \emph{agentic relevance}, by prioritizing examples that likely require tools to acquire, transform, or revisit visual evidence rather than relying on direct one-shot perception. To enforce the last criterion efficiently, we apply a hierarchical filtering pipeline that first identifies tool-relevant instances and then removes samples that are either trivially solvable without tools or likely to be noisy. This produces a higher-quality set of training instances for subsequent trajectory synthesis. Detailed data source and filtering procedures are deferred to Sec.~S1 of the supplementary material.

\subsubsection{Trajectory Generation.}
Given the curated candidate pool, we synthesize trajectories in two complementary phases: a plan-then-execute pipeline that generates grounded single-tool and compositional multi-tool trajectories, and a tool-context reinstantiation step that diversifies tool interfaces and tool-use contexts to improve adaptive tool-use capability.

\begin{figure*}[t]
 \centering

 \begin{subfigure}[t]{0.72\textwidth}
 \centering
 \includegraphics[width=\textwidth]{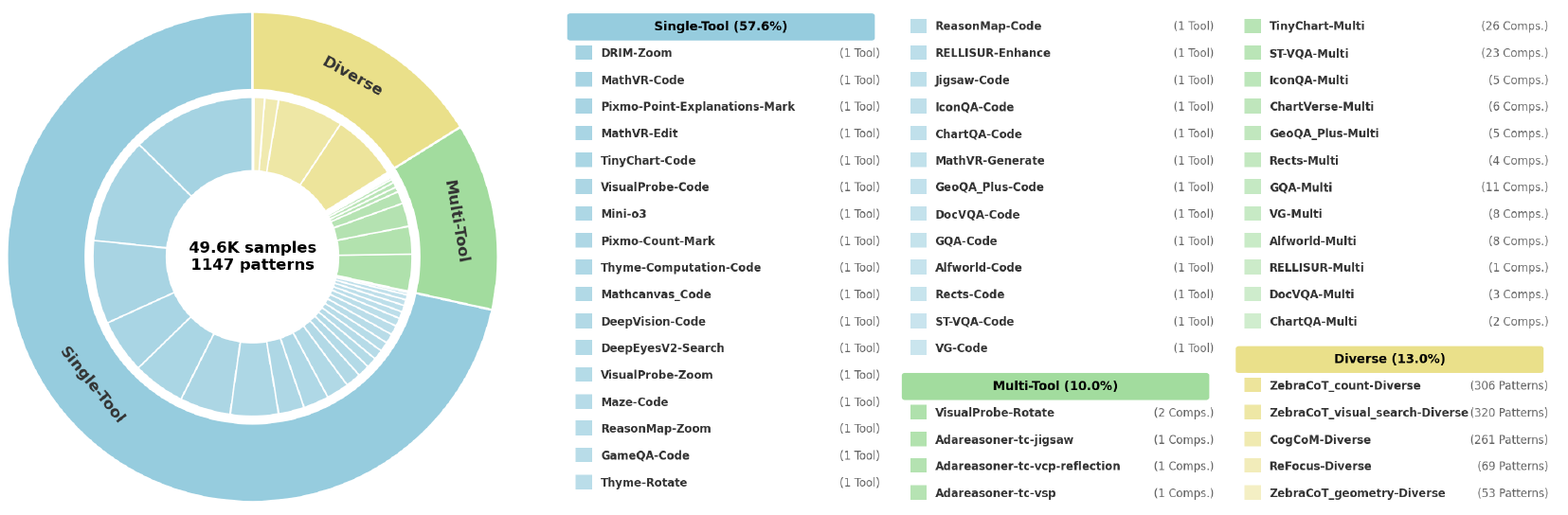}
 \caption{Composition of the curated SFT trajectory bank.}
 \label{fig:fig3_left}
 \end{subfigure}
 \hfill
 \rule{0.5pt}{4.0cm}
 \hfill
 \begin{subfigure}[t]{0.25\textwidth}
 \centering
 \includegraphics[width=\textwidth]{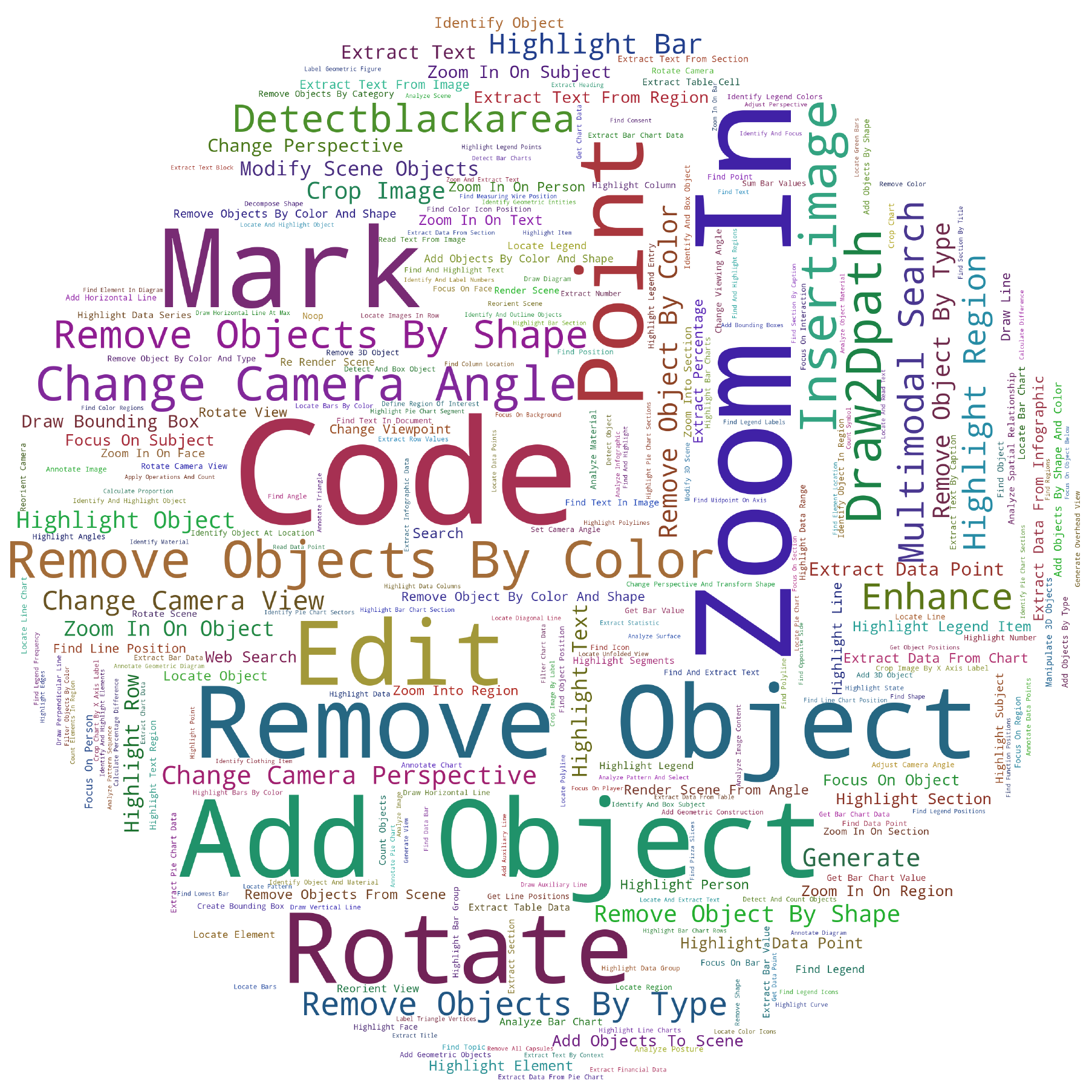}
 \caption{Tool and pattern diversity.}
 \label{fig:fig3_right}
 \end{subfigure}

 \caption{\textbf{Composition and diversity of the synthesized SFT trajectory bank.}
(a) Breakdown of the curated trajectories into single-tool, multi-tool, and diverse-tool-context data, together with their sample and pattern distributions.
(b) Word cloud of tool and operation patterns covered in the trajectory bank.}
 \label{fig:fig3}
  \vspace{-0.4cm}
\end{figure*}

\noindent\textit{Plan-then-Execute Trajectory Synthesis.}
This pipeline separates high-level tool planning from grounded execution to support both single-tool grounding and compositional multi-tool use. In the planning phase, a large thinking VLM first generates an ordered tool-use plan
\begin{equation}
\pi = (t^{\text{plan}}_1, \ldots, t^{\text{plan}}_M) \sim p_{\text{plan}}(\cdot \mid q, \hat{y}, v, T),
\end{equation}
where $\hat{y}$ is the answer annotation. In the execution phase, a non-thinking model follows the plan step by step in the tool environment. At step $k$, given the interaction history
\begin{equation}
h_k = (q, v, s_1, a_1, o_1, \ldots, s_{k-1}, a_{k-1}, o_{k-1}),
\end{equation}
and the planned tool $t^{\text{plan}}_k$, the executor predicts the reasoning state and tool arguments:
\begin{equation}
\begin{aligned}
(s_k, u_k) &\sim p_{\mathrm{exec}}(\cdot \mid h_k, t^{\mathrm{plan}}_k), \\
a_k = (t^{\mathrm{plan}}_k&, u_k), \quad o_k = \mathrm{Env}(a_k).
\end{aligned}
\end{equation}
Generation stops when the plan is exhausted or no valid tool call is produced, and we categorize the resulting trajectories as single- or multi-tool according to the number of tool calls. This decomposition offers two advantages. First, using the answer annotation $\hat{y}$ only in the planning stage leads to higher-quality tool-use plans while avoiding answer leakage into execution. Second, executing tools under a planned sequence provides a controlled interaction context, improving the robustness of compositional tool use by grounding each step in tool-returned observations. 

\noindent\textit{Diverse Tool-Contexts Reinstantiation.}
To enable the model to adapt to diverse and potentially unseen tool contexts, we augment the synthesized trajectories with varied tool interfaces while preserving the underlying visual operations. For each step $k$, conditioned on the pre-action visual state $o_{k-1}$ (with $o_0\!:=\!v$), the post-action state $o_k$, the reasoning $s_k$, and the current tool pool $\mathcal{T}$, we jointly sample a reinstantiated tool schema and its arguments:
\begin{equation}
(\tilde{t}_k,\tilde{u}_k) \sim p_{\text{ctx}}(\cdot \mid o_{k-1}, o_k, s_k, \mathcal{T}),
\end{equation}
In practice, $p_{\text{ctx}}$ is realized by prompting a strong VLM to inspect the visual transition and either match it to an existing tool in $\mathcal{T}$ or define a new schema when no suitable candidate exists. The observation $o_k$ from the original execution is kept intact, so $\tilde{a}_k=(\tilde{t}_k,\tilde{u}_k)$ realizes the same underlying visual operation through a different interface. This exposes the model to multiple interface realizations for the same underlying operation, encouraging it to adapt tool calls based on visual affordances rather than memorize fixed invocation patterns. To maximize the diversity of synthesized tools, we build these trajectories from implicit reasoning datasets such as Zebra-CoT~\cite{li2025zebra} and Monet~\cite{wang2025monet}. Representative synthesized tool schemas and the full prompt are shown in Sec.~S3 of the supplementary material.

\noindent\textbf{Post-processing and Filtering.}
After trajectory synthesis, we post-process the data by filtering out answer-inconsistent, format-invalid, or inefficient trajectories, and introduce distractor tools into the candidate pool during training for robustness. Detailed procedures are provided in Sec.~S4 of the supplementary material.

As summarized in Fig.~\ref{fig:fig3}, the resulting SFT trajectory bank covers a broad spectrum of tool-use trajectories, ranging from foundational single-tool grounding to multi-step composition and diverse tool contexts. Specifically, the single-tool and multi-tool subsets involve six executable tools (e.g., zoom in), whereas the diverse-context subset augments them with over 1{,}000 mock but valid tool interfaces, substantially increasing interface diversity for adaptive tool-use training. In addition, we include a small number of no-tool reasoning samples from ChartVerse~\cite{liu2026chartverse} to preserve direct answering capability.

\subsection{Two-Stage Training}
\label{sec:training}

We train VC-Tooler in two stages: a supervised cold start on the synthesized trajectory bank to establish the three tool-use capabilities, followed by agentic RL that refines the resulting policy through interaction with the tool environment.

\subsubsection{Cold-Start SFT.}
\label{sec:sft}

In the first stage, we fine-tune the base VLM on the SFT trajectory bank of Sec.~\ref{sec:synthesis} with a standard token-level cross-entropy loss over the assistant turns of each trajectory. Because the bank already covers single-tool grounding, multi-tool composition, and diverse tool contexts, this stage equips the model with a broad set of tool-use behaviors before any environment interaction.

\subsubsection{Agentic RL.}
\label{sec:rl}

Building on this cold-started policy, we further refine VC-Tooler with GRPO~\cite{shao2024deepseekmath} using a tool reward that encourages the model to sharpen its compositional and adaptive tool use.

Concretely, the tool-reward module is a lightweight critic that inspects a rollout trajectory $\tau$ and returns a scalar reward $R_{\mathrm{tool}}(\tau)\in[0,1]$. Rather than rewarding tool use per se, it measures a small set of tool-grounded behaviors, primarily whether the model incorporates returned observations into subsequent reasoning and avoids redundant or non-progressing calls. We use $R_{\mathrm{tool}}$ as a shaping term during RL, and defer the detailed criteria and prompt to Sec.~S5 of the supplementary material.
  
\begin{table*}[t!]
\centering
\caption{\textbf{Overall performance comparison on general-purpose and agentic benchmarks.} The best results are in \textbf{bold}, and the second-best are \underline{underlined}. }
\label{tab:main_results}
\resizebox{0.9\textwidth}{!}{
\begin{tabular}{l | ccccccc c c}
\toprule
\multirow{3}{*}{\textbf{Model}}
& \multicolumn{6}{c}{\textbf{General-Purpose Benchmarks}}
& \multicolumn{2}{c}{\textbf{Agentic Benchmarks}} \\
\cmidrule(lr){2-7} \cmidrule(lr){8-9}
& \multirow{2}{*}{\textbf{V*}}
& \textbf{HRBench} & \textbf{HRBench}
& \multicolumn{2}{c}{\textbf{CharXiv}}
& \multirow{2}{*}{\textbf{MME-RW}}
& \multirow{2}{*}{\textbf{VTC-Bench}}
& \multirow{2}{*}{\textbf{TIR-Bench}} \\
& & \textbf{4K} & \textbf{8K} & \textbf{DQ} & \textbf{RQ} & & & \\
\midrule
\multicolumn{9}{c}{\textit{General-Purpose MLLMs}} \\
\midrule
Qwen3-VL-8B~\shortcite{yang2025qwen3} & 90.1 & 82.3 & 78.0 & 83.0 & 46.4 & 61.9 & 28.7 & 18.4 \\
LLaVA-OV~\shortcite{li2024llava} & 75.4 & 63.0 & 59.8 & - & - & 57.4 & -- & -- \\
InternVL3-8B~\shortcite{zhu2025internvl3} & 81.2 & 70.0 & 69.3 & 73.6 & 37.6 & -- & -- & 16.9 \\
o4-mini~\shortcite{openai2025thinking} & 94.6 & - & - & 94.3 & 72.0 & -- & 33.7 & 37.5 \\
\midrule
\multicolumn{9}{c}{\textit{Agentic Reasoning Models}} \\
\midrule
DeepEyes~\shortcite{zheng2025deepeyes} & 85.6 & 75.1 & 72.6 & -- & -- & -- & 29.3 & 17.3 \\
Thyme~\shortcite{zhang2025thyme} & 82.2 & 77.0 & 72.0 & 66.9 & 44.5 & \underline{64.8} & 28.8 & 14.8 \\
DeepEyes-V2~\shortcite{hong2025deepeyesv2} & 81.8 & 77.9 & 73.8 & 78.6 & 48.9 & \underline{64.9} & \underline{30.0} & -- \\
AdaReasoner~\shortcite{song2026adareasoner} & 81.7 & 73.6 & 70.1 & 58.4 & 38.2 & 61.3 & 29.3 & -- \\
PyVision-RL~\shortcite{zhao2026pyvisionrl} & 88.7 & 78.1 & 74.3 & -- & -- & -- & -- & 19.8 \\
CodaDance~\shortcite{Song2025codedance} & 84.8 & 75.2 & 72.3 & -- & 44.1 & -- & -- & -- \\
\midrule
\textbf{VC-Tooler-SFT}
& \underline{92.1} & \underline{82.9} & \underline{78.6}
& \underline{83.8} & \underline{51.7} & 62.9
& 27.8 & \underline{20.2} \\
\textbf{VC-Tooler-RL}
& \textbf{95.8} & \textbf{83.9} & \textbf{83.8}
& \textbf{84.0} & \textbf{54.1} & \textbf{69.5}
& \textbf{35.3} & \textbf{21.0} \\
$\Delta$ (vs Qwen3-VL-8B)
& +5.7 & +1.6 & +5.8 & +1.0 & +7.7 & +7.6 & +6.6 & +2.6 \\
\bottomrule
\end{tabular}
}
 \vspace{-0.4cm}
\end{table*}

We combine this tool reward with standard accuracy and format rewards for answer correctness and output validity:
\begin{equation}
R_{total}=0.8 R_{\mathrm{acc}}+ 0.2 R_{\mathrm{tool}}+0.2 R_{\mathrm{fmt}},
\label{eq:total_reward}
\end{equation}
where \(R_{\mathrm{acc}}\) is the primary reward evaluating whether the final answer is correct, \(R_{\mathrm{fmt}}\) is a standard format-validity term, and \(R_{\mathrm{tool}}\) provides auxiliary shaping toward compositional and adaptive tool use.

For RL data, we collect training instances from existing multimodal reasoning and tool-use RL data sources, including ChartVerse~\cite{liu2026chartverse}, DeepEyes~\cite{zheng2025deepeyes}, DRIM~\cite{yang2025deep}, VisualProbe~\cite{lai2025mini}. To improve optimization stability, we further filter out unsalvageable instances before policy training. Detailed dataset composition and filtering criteria are provided in Sec.~S6 of the supplementary material.

\section{Experiment}
\subsection{Implementation Details}
\label{sec:impl_details}
Our training framework consists of two stages: cold-start SFT and agentic reinforcement learning. The base model is Qwen3-VL-8B~\cite{yang2025qwen3}. In the SFT stage, we train the base model for 3 epochs with a batch size of 32, a learning rate of 1e-5, AdamW~\cite{loshchilov2017decoupled} optimizer, and cosine learning rate decay. For RL, we adopt GRPO with a batch size of 128, 8 rollouts per prompt, and a learning rate of 1e-6. We set the KL coefficient to 0.0, the maximum number of interaction turns during training to 5, and set the maximum response length to 32,768 tokens, and allow up to 10 turns during inference to leave room for longer tool-use trajectories.
We use ms-swift~\cite{zhao2025swift} and verl~\cite{sheng2025hybridflow} as the training frameworks for SFT and RL, respectively.

\subsection{Evaluation Details}
We evaluate VC-Tooler on two categories of benchmarks. General-purpose benchmarks include V*~\cite{wu2024v}, HRBench-4K/8K~\cite{wang2025divide}, CharXiv~\cite{wang2024charxiv}, and MME-RealWorld~\cite{zhang2025mme}, measuring broad visual understanding capabilities. Agentic benchmarks include VTC-Bench~\cite{zhu2026vtc} and TIR-Bench~\cite{li2025tir}, targeting compositional agentic visual reasoning. We compare against both open-source general-purpose MLLMs and agentic reasoning models. For inference, we combine Qwen-Agent with vLLM and use lmms-eval as the evaluation framework.

\noindent\textbf{Tool pool in evaluation.} We evaluate VC-Tooler with a diverse tool pool consisting of \emph{seen} tools and \emph{unseen} tools. The seen tools are the six tools available during training (\texttt{zoom}, \texttt{rotate}, \texttt{enhance}, \texttt{code}, \texttt{mark}, \texttt{edit}), whereas the unseen tools are eleven additional tools (e.g., \texttt{flip}, \texttt{brightness}, \texttt{histeq}, \texttt{denoise}, \texttt{sharpen}, \texttt{inpaint}, \texttt{morph}) that the model sees only through natural-language schemas at inference. Code is executed in a local sandbox, and the edit tool calls an online image-editing model (Qwen-Image~\cite{wu2025qwen}). For baselines, we report results either from the original papers or from our re-evaluation under each method’s native tool setting.

\begin{figure}[t]
\centering
\includegraphics[width=\columnwidth]{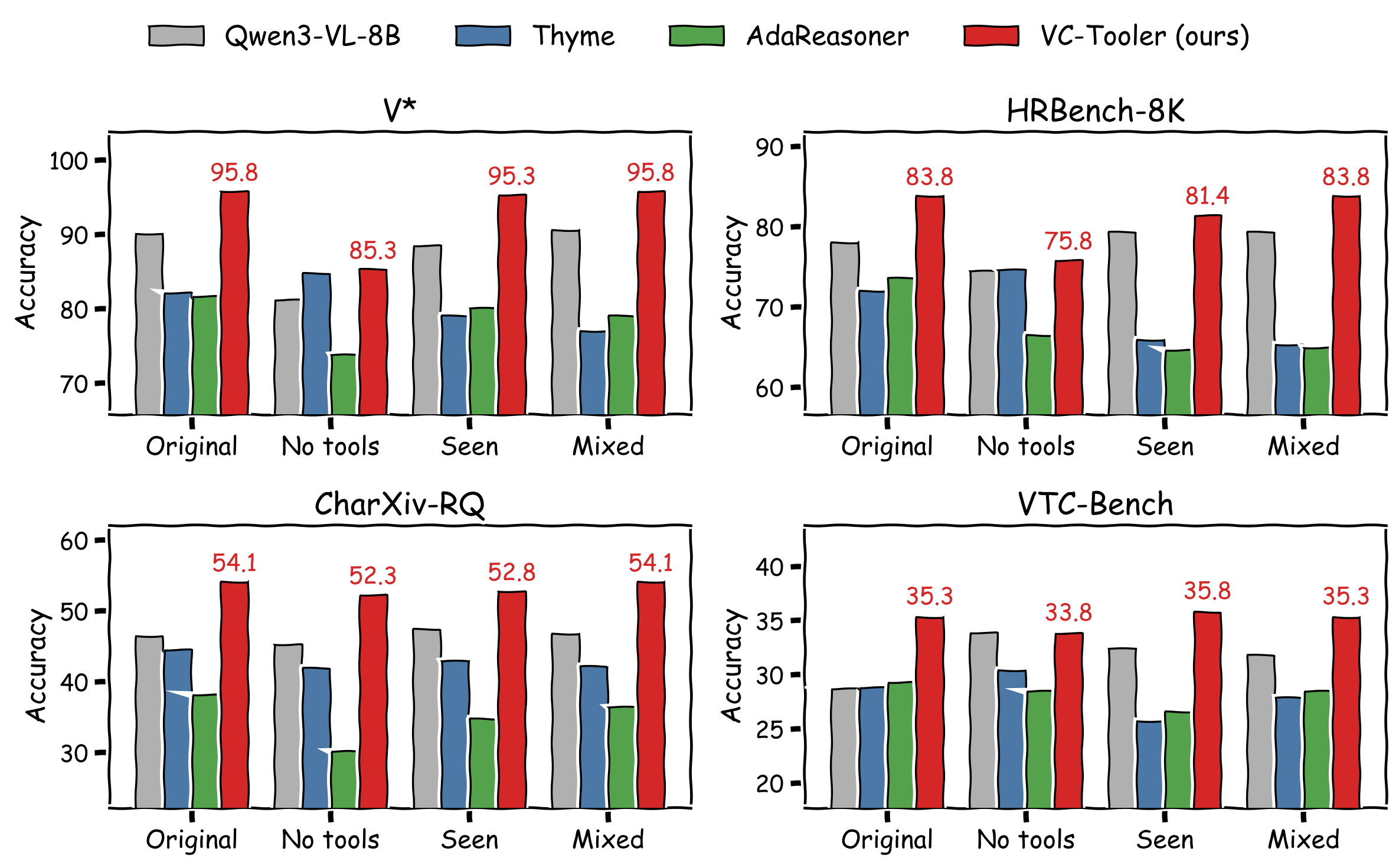}
\caption{\textbf{Generalization across tool availability.} Per-benchmark accuracy of all methods under four tool configurations (\emph{Original}, \emph{No tools}, \emph{Seen}, \emph{Mixed}).}
\label{fig:tool_generalization}
 \vspace{-0.4cm}
\end{figure}

\begin{figure*}[t]
 \centering
 \includegraphics[width=0.95\textwidth]{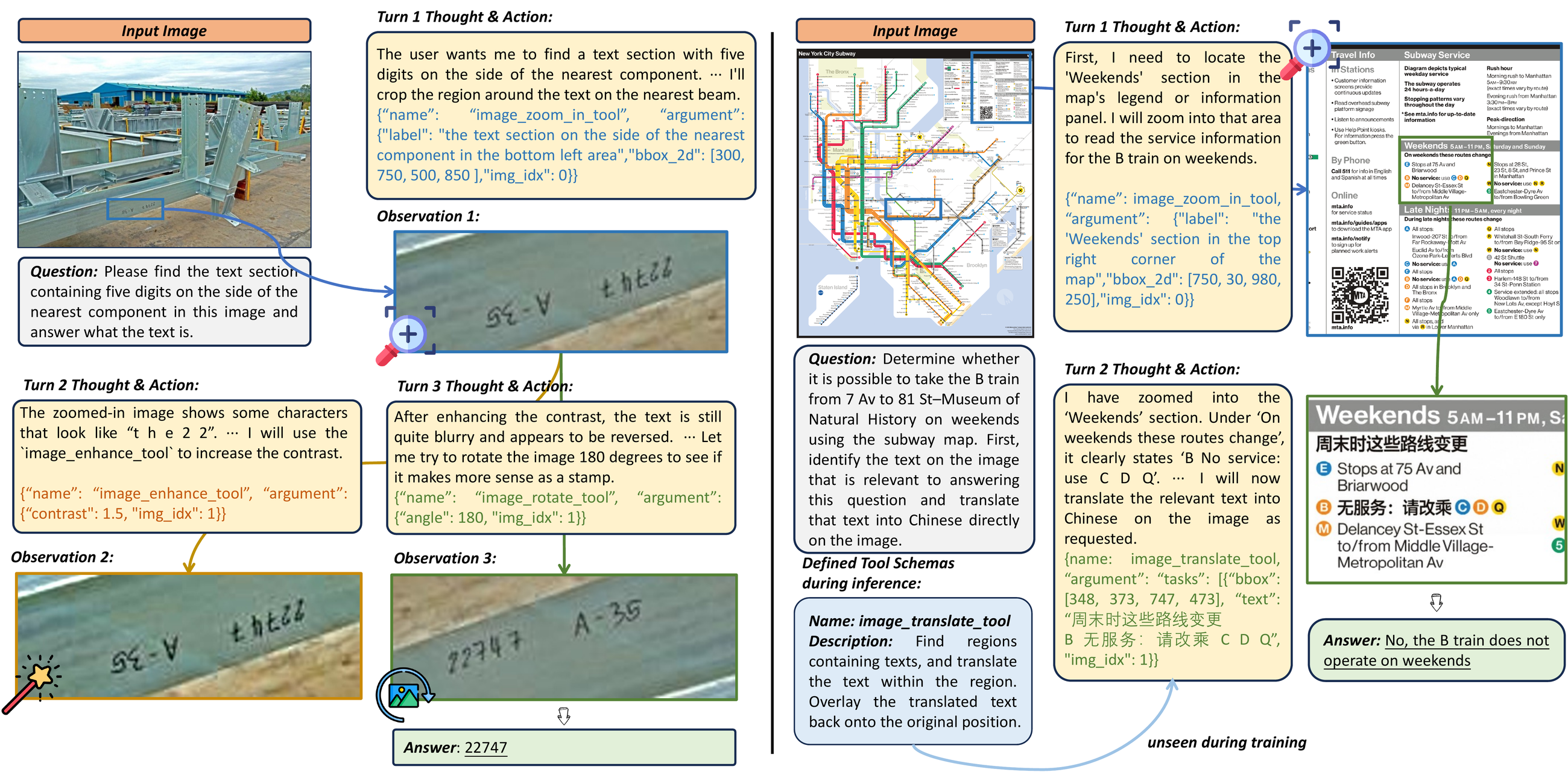}
 \caption{\textbf{Case studies of compositional and adaptive visual tool use of VC-Tooler.}}
 \label{fig:case-study}
  \vspace{-0.4cm}
\end{figure*}

\subsection{Main Results}
Table~\ref{tab:main_results} reports the performance of VC-Tooler on both general-purpose and agentic benchmarks. We highlight three observations.

\noindent\textbf{VC-Tooler performs strongly across both general-purpose and agentic benchmarks.} On general-purpose benchmarks, VC-Tooler-RL attains 95.8 on V*, 83.8 on HRBench-8K, and 69.5 on MME-RealWorld, outperforming prior open-source agentic baselines and remaining competitive with strong proprietary models. It also improves over prior open-source methods on agentic reasoning benchmarks. Taken together, these results suggest that the proposed training framework is useful not only for specialized tool-use tasks, but more broadly for visual reasoning settings in which tools can help acquire, revisit, or refine visual evidence.

\noindent\textbf{The gains are especially clear on benchmarks that require multi-step tool use.} On VTC-Bench, VC-Tooler-RL reaches 35.3, exceeding the best reported open-source baseline by 5.3 points. This benchmark requires the model to reason over tool-returned observations across multiple steps rather than rely on a single grounded tool call. The strong improvement therefore aligns with our central objective of training visual tool use as a compositional capability, instead of limiting the model to isolated tool invocation patterns.

\noindent\textbf{VC-Tooler remains robust under richer and partially novel tool settings.}
 We further evaluate all methods under four configurations: \emph{Original}, which keeps each method's native tool environment; \emph{No tools}, which removes tool use entirely; \emph{Seen}, which provides only the seen tool pool; and \emph{Mixed}, which augments the seen pool with unseen tools. As shown in Fig.~\ref{fig:tool_generalization}, prior methods generally do not benefit from expanded tool spaces and can even degrade when additional tools are introduced. In contrast, VC-Tooler matches or exceeds its original-setting performance under the seen and mixed pools, despite not observing the added tools during training. This robustness supports our main claim that effective visual tool use should be learned as a capability conditioned on tool schemas and interaction context, rather than as a fixed set of familiar calling patterns.

\subsection{Ablation Study}

\begin{figure}[t!]
 \centering
 \includegraphics[width=\columnwidth]{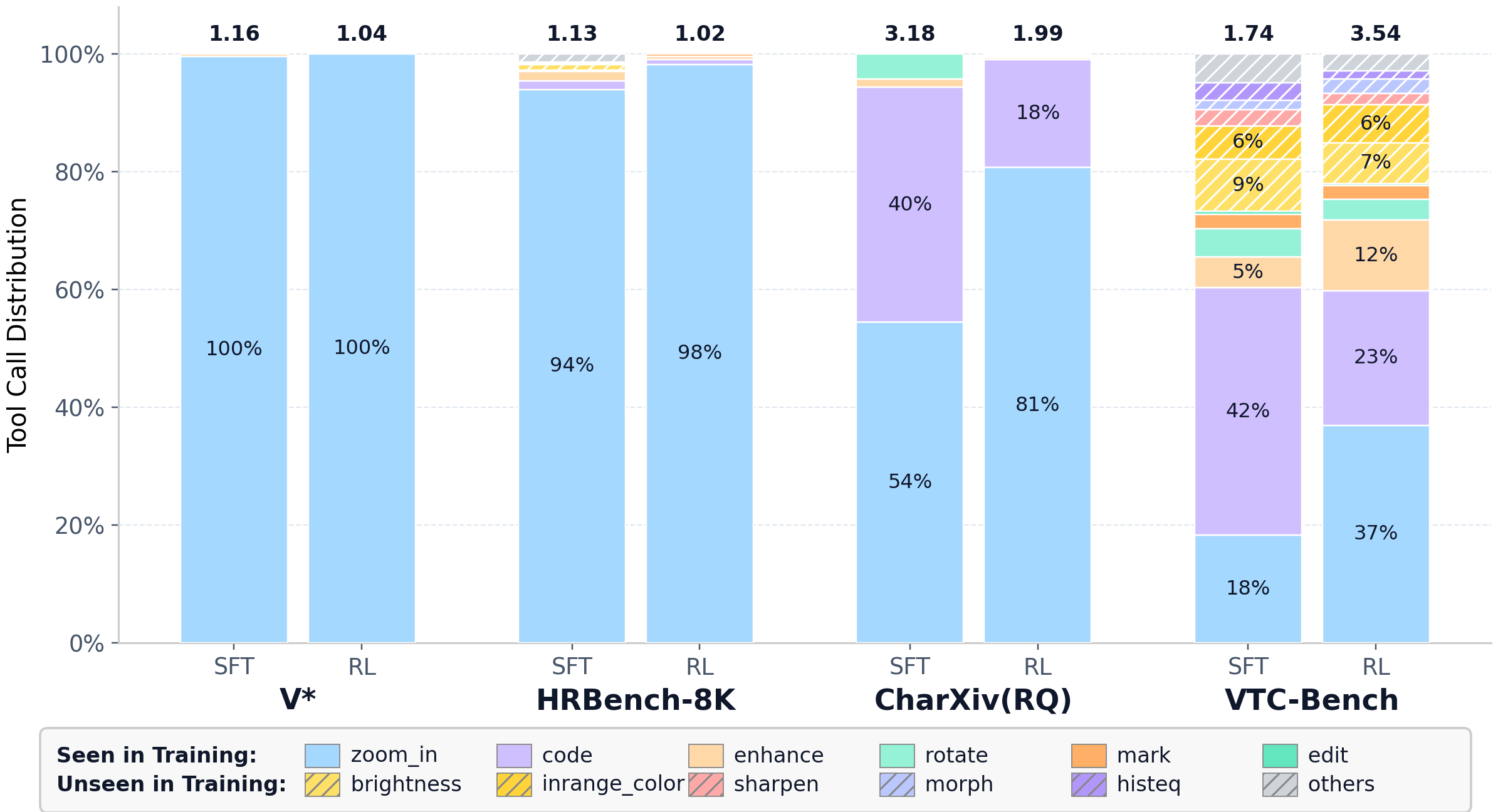}
 \caption{\textbf{Tool-call distribution before and after RL.} Each bar shows the proportion of invoked tools, with solid colors denoting tools seen during training and hatched segments denoting novel tools. The number on top of each bar indicates the average number of tool calls per benchmark.}
 \label{fig:sft_rl_comparison}
  \vspace{-0.2cm}
\end{figure}

\subsubsection{Effect of SFT Data Composition}
To understand how different types of supervised data contribute to VC-Tooler, we ablate the synthesized three trajectory levels: foundational single-tool use~(ST), multi-tool composition~(MT), and diverse tool contexts~(DTC). Table~\ref{tab:sft_rl_ablation} shows how different combinations progressively shape tool-use capability.

We first anchor the comparison with two references. Directly fine-tuning on the same data with tool calls removed (\emph{Direct SFT}) does not help and even falls below the untrained \emph{Base} model on V* (88.5 vs.\ 90.1) and HRBench-8K (75.9 vs.\ 78.0), whereas every tool-augmented composition clearly surpasses both. This indicates that the improvements stem from learning to use tools rather than from supervised fine-tuning on this data alone. Using only foundational single-tool data yields strong SFT performance, suggesting that it provides a relatively easy and stable initialization for visual-grounded invocation. 
By contrast, adding MT or DTC data does not always improve SFT immediately, and can even hurt performance under pure imitation learning. We attribute this to the greater difficulty of learning from longer-horizon interactions and more heterogeneous tool schemas.

\begin{table}[t!]
\centering
\small
\setlength{\tabcolsep}{4pt}
\caption{\textbf{Effect of different Stage~I data compositions.} ST denotes foundational single-tool use, MT multi-tool composition, and DTC diverse tool contexts. We also report two references: the untrained \emph{Base} model and \emph{Direct SFT}, trained on the same data with tool calls removed. We report performance after SFT and after subsequent RL on V*, HRBench-8K, CharXiv(RQ), and VTC-Bench.}
\label{tab:sft_rl_ablation}
\begin{tabular}{l|cccc}
\toprule
\textbf{SFT Data} & \textbf{V*} & \textbf{HR-8K} & \textbf{CX-RQ} & \textbf{VTC} \\
\midrule
Base (no SFT) & 90.1 & 78.0 & 46.4 & 28.7 \\
Direct SFT (w/o tools) & 88.5 & 75.9 & 49.3 & 32.8 \\
\midrule
\multicolumn{5}{c}{\textbf{After SFT}} \\
\midrule
ST & 91.6 & 77.8 & 55.2 & 33.6 \\
ST\,+\,MT & 89.0 & 75.9 & 51.7 & 31.6 \\
ST\,+\,DTC & 91.1 & 79.5 & 55.3 & 29.7 \\
ST\,+\,MT\,+\,DTC & 92.1 & 78.6 & 51.7 & 30.4 \\
\midrule
\multicolumn{5}{c}{\textbf{After RL}} \\
\midrule
ST & 94.2 & 80.4 & 55.3 & 34.0 \\
ST\,+\,MT & 90.6 & 80.6 & 52.8 & 33.7 \\
ST\,+\,DTC & 94.8 & 83.1 & 53.3 & 34.4 \\
ST\,+\,MT\,+\,DTC & 95.8 & 83.8 & 54.1 & 35.3 \\
\bottomrule
\end{tabular}
 \vspace{-0.4cm}
\end{table}

This trend reverses after RL. Compared with ST-only, adding DTC improves post-RL performance on HRBench-8K from 80.4 to 83.1, while the full combination of ST, MT, and DTC yields the strongest overall policy, reaching 95.8 on V*, 83.8 on HRBench-8K, and 35.3 on VTC-Bench. These results suggest that SFT data should not be judged only by immediate SFT accuracy: data that is harder to imitate can still provide useful structural priors for downstream policy optimization. Overall, ST supports reliable tool invocation, MT broadens compositional exploration, and DTC improves adaptation across schemas and contexts; together, they produce the best final policy.

\subsubsection{Effect of RL Design}
We next study how different RL designs affect visual tool use. In particular, we compare three settings: (1) standard outcome and format reward without any explicit tool-use signal, (2) an additional tool-existence reward that encourages calling tools, and (3) our tool reward that evaluates whether tool feedback is actually incorporated into subsequent reasoning.

As shown in Table~\ref{tab:rl_reward_ablation}, simply rewarding tool calls does not improve over the standard reward and instead hurts performance across benchmarks, suggesting that encouraging tool invocation alone can bias the model toward superficial action patterns. By contrast, adding tool reward consistently performs best, with gains of up to 1.8 points over the standard reward, indicating that it provides complementary supervision beyond sparse outcome signals by rewarding the effective use of tool feedback in subsequent reasoning.

\subsubsection{Analysis of Learned Tool-Use Behavior}
To better understand the policy learned by VC-Tooler, we analyze its tool-use patterns across benchmarks. Fig.~\ref{fig:sft_rl_comparison} shows that the model does not adopt a uniform strategy of calling more tools. Instead, its behavior varies with task demands.

On relatively simple benchmarks such as V* and HRBench-8K, tool use remains highly concentrated on zoom, with a small average number of calls (e.g., $1.16\xrightarrow{}1.04$ on V*). This suggests that VC-Tooler learns a selective policy for tasks where localized inspection is usually sufficient. On more agentic benchmarks, however, the learned behavior becomes substantially richer. On CharXiv(RQ), the model shifts from heavy reliance on code to more selective visual inspection, while on VTC-Bench it uses a wider range of tools and longer interaction trajectories. The latter also shows greater use of unseen tools, indicating that the model can incorporate newly provided tool schemas when the task benefits from additional operations.

Overall, these patterns suggest that VC-Tooler learns task-dependent tool-use strategies rather than a single fixed interaction pattern. It stays economical on simpler tasks, while expanding tool diversity and interaction depth only when multi-step evidence acquisition is necessary. This behavior is consistent with our goal of learning visual tool use as a compositional and adaptive capability.

\begin{table}[t]
\centering
\small
\caption{\textbf{Ablation on RL reward design.} Standard reward (Std) contains basic answer and format reward. Tool-existence reward (TE) simply encourages tool invocation, while the tool reward (Tool) encourages effective use of tool-returned observations.}
\label{tab:rl_reward_ablation}
\begin{tabular}{c c c|cccc}
\toprule
\textbf{Std} & \textbf{TE} & \textbf{Tool} & \textbf{V*} & \textbf{HR-8K} & \textbf{CX-RQ} & \textbf{VTC} \\
\midrule
\checkmark & & & 94.7 & 82.0 & 52.9 & 33.4 \\
\checkmark & \checkmark & & 93.2 & 79.0 & 52.1 & 33.1 \\
\checkmark & & \checkmark & \textbf{95.8} & \textbf{83.8} & \textbf{54.1} & 35.3 \\
\bottomrule
\end{tabular}
 \vspace{-0.4cm}
\end{table}

\section{Conclusion}
We introduced VC-Tooler, an agentic multimodal model that learns visual tool use as a compositional and adaptive capability. By combining hierarchical trajectory synthesis with supervised cold start and tool-aware reinforcement learning, VC-Tooler achieves strong results on both general-purpose and agentic benchmarks, especially on multi-step tool-use tasks, and remains robust under richer and partially novel tool settings. These results suggest that visual tool use can be learned as compositional and adaptive capability rather than a fixed set of calling patterns.

\section{Limitations}
Our study is limited in scale: trajectory synthesis and RL training remain modest relative to the complexity of visual tool use, and we still observe gains from further scaling. Current experiments also focus on bounded tool spaces and short interaction horizons, leaving larger-scale training and more open-ended settings to future work.

\bibliography{custom}

% =====================================================================
% Supplementary Material (merged after the main text).
% The main paper keeps default section/figure/table numbering; from here
% on we switch to an "S"-prefixed style so that supplementary sections,
% figures, and tables match the "Sec.~SN" references made in the main text
% and cannot collide with the main-paper numbering.
% =====================================================================
\renewcommand{\thesection}{S\arabic{section}}
\renewcommand{\thesubsection}{S\arabic{section}.\arabic{subsection}}
\renewcommand{\thefigure}{S\arabic{figure}}
\renewcommand{\thetable}{S\arabic{table}}
\setcounter{section}{0}
\setcounter{figure}{0}
\setcounter{table}{0}

% Start the supplementary material on a fresh page with a full-width title
% banner. \twocolumn[...] issues the page break and spans the header across
% both columns before resuming the two-column body.
\twocolumn[
 \begin{center}
  {\LARGE\bfseries Supplementary Material}\par
  \vspace{0.6em}
  {\large VC-Tooler: Learning Compositional and Adaptive Visual Tool Use}\par
 \end{center}
 \vspace{1.5em}
]

% =====================================================================
% Supplementary appendix, reorganized to follow the main-paper order:
% Trajectory Synthesis (S1--S3) -> Training (S4--S7) -> Evaluation (S8--S10)
% S-prefixed numbering is set in supplementary.tex.
% =====================================================================

\section{Detailed Data Selection Pipeline}
\label{app:data_selection}

We construct candidate samples for trajectory synthesis from large-scale multimodal datasets, with the goal of retaining instances that are diverse, verifiable, tool-relevant, and sufficiently challenging for agentic visual reasoning. In practice, we source candidates from broad-coverage multimodal corpora, including LLaVA-OneVision and other public datasets spanning chart, document, STEM, logic, and general visual reasoning domains. This broad sourcing strategy helps ensure that the resulting trajectory bank covers heterogeneous visual scenes, question formats, and tool-use patterns.

\paragraph{Data sources.}
For clarity, Table~\ref{tab:dataset_classification} groups the datasets used during candidate construction by task category. This categorization is intended only to summarize coverage; our filtering and trajectory synthesis procedures are applied at the sample level rather than the dataset level.

\begin{table}[t]
\centering
\caption{Dataset classification by category used in candidate construction.}
\label{tab:dataset_classification}
\small
\renewcommand{\arraystretch}{1.12}
\setlength{\tabcolsep}{4pt}
\begin{tabular}{p{0.21\columnwidth} p{0.67\columnwidth}}
\toprule
\textbf{Category} & \textbf{Datasets} \\
\midrule
\textbf{Agentic} &
Thyme, DRIM, AdaReasoner, Mini-o3, \newline
VisualProbe, ReasonMap \\

\textbf{Chart/Doc} &
TinyChart, ChartQA, ChartVerse, DocVQA \\

\textbf{STEM} &
GeoQA, MathVR, MathCanvas \\

\textbf{General} &
GQA, ST\_VQA, VG, Rects, Alfworld, \newline
Pixmo, RELLISUR \\

\textbf{Implicit} &
ZebraCOT, Monet \\

\textbf{Logic} &
DeepVision, IconQA, GameQA \\
\bottomrule
\end{tabular}
\end{table}

\paragraph{Tool-relevance filtering.}
We first apply a prompt-based judge to identify instances that are likely to require external tools. Samples predicted to be directly solvable by simple one-shot perception are discarded at this stage. This step allows us to prioritize examples that require acquiring, refining, or transforming visual evidence, rather than examples answerable from an immediate glance.

We use Qwen3-VL-32B~\citep{yang2025qwen3} as the tool-relevance judge. The model receives the input image and question, and must output a binary decision indicating whether a tool call is needed. The exact prompt is shown in Prompt~\ref{prompt:tool_need_routing_prompt}.

\begin{prompt*}[t]
\centering
\begin{tcolorbox}[
 width=\textwidth,
 colback=black!3!white,
 colframe=black!75!white,
 rounded corners,
 arc=1mm,
 boxrule=0.2mm,
 title=Prompt for Tool-relevance filtering
]
\footnotesize
You are an AI task router. Your primary and ONLY job is to analyze a user's question about an image or video and determine if you are capable of answering it directly.

\textbf{Core Principle:}

You must recognize your own limitations. If a question requires any capability beyond basic visual perception (e.g., calculation, external knowledge, image manipulation), you are INCAPABLE of answering it by yourself. In such cases, a tool call is mandatory. Do not attempt to guess or hallucinate an answer.

\textbf{Available Tools:}

\texttt{image\_zoom\_in\_tool}: Zooms in on a specific region of the image to see small details.\\
\texttt{image\_rotate\_tool}: Rotates the image if it is oriented incorrectly.\\
\texttt{image\_enhance\_tool}: Improves image quality, contrast, or brightness for better visibility.\\
\texttt{image\_code\_tool}: Runs Python code for precise numerical computation, geometric plotting, solving mathematical equations, or generating coordinate-based visual models from image data.\\
\texttt{multimodal\_search}: Performs a web search using text or an image to identify specific entities (landmarks, celebrities, products) or fetch real-time knowledge.\\
\texttt{image\_mark\_tool}: Annotates the image by drawing points, bounding boxes, or arrows to highlight specific locations or count dense objects.\\
\texttt{image\_edit\_tool}: Modifies specific elements of an existing image.\\
\texttt{others\_tool}: A long-tail tool for specialized or niche requests that do not fit into the categories above but require external processing (e.g., professional domain analysis, specialized file parsing, or complex multi-step reasoning not covered by standard tools).

\textbf{Decision-Making Guidelines:}

Apply the Core Principle using these examples. Assume a tool is needed unless the answer is trivially obvious from a glance.

\textbf{Tool Needed (YES):}
Computational/Geometric: The question requires calculating distances, areas, or plotting geometric shapes. $\rightarrow$ \texttt{image\_code\_tool}.\\
External Knowledge: Requires identifying entities (landmarks, people) or facts not present in the image. $\rightarrow$ \texttt{multimodal\_search}.\\
Visual Manipulation: Requires marking, zooming, rotating, or enhancing the image. $\rightarrow$ \texttt{image\_mark\_tool}, \texttt{image\_zoom\_in\_tool}, etc.\\
Content Modification: Requires generating or editing visual content. $\rightarrow$ \texttt{image\_edit\_tool}.\\
Specialized/Long-tail: The task is complex and does not fit the tools above but still requires external processing. $\rightarrow$ \texttt{others\_tool}.

\textbf{No Tool Needed (NO):}
A tool is not needed ONLY for questions answerable by direct, simple visual perception. This includes identifying primary colors, naming globally obvious objects (e.g., ``a car,'' ``a tree''), describing the overall scene in general terms, or counting a few, very clear objects ($<5$).

\textbf{Output Format:}

You must strictly output ONLY one word (\texttt{YES} / \texttt{NO}). \texttt{YES} for tool\_call needed, while \texttt{NO} for no tool needed. DO NOT provide any explanation.

\textbf{Your Task:}

\texttt{[Input]: <image>} \\
\texttt{[Question]: \{question\}}

Based on the rules above, analyze the input and provide the output.
\end{tcolorbox}
\caption{Prompt used for tool-relevance filtering.}
\label{prompt:tool_need_routing_prompt}
\end{prompt*}

\paragraph{Difficulty filtering.}
Among the retained tool-relevant samples, we further estimate problem difficulty using repeated sampling from a baseline model, Qwen3-VL-8B-Instruct~\citep{yang2025qwen3}. Specifically, we sample 8 responses per instance. If a sample is solved in at least 4 out of 8 attempts, we regard it as trivially easy and remove it from the synthesis pool. This criterion removes instances that do not meaningfully require robust tool-grounded reasoning.

For samples unsolved in all 8 attempts, we further distinguish between genuinely difficult instances and potentially noisy or ill-posed ones by re-evaluating them with two stronger judges: Qwen3-VL-235B-A22B-Instruct and Qwen3-VL-8B-Thinking~\citep{yang2025qwen3}. Instances that remain unsolved by all judges are discarded as likely ambiguous, noisy, or annotation-misaligned. In contrast, instances solved by at least one stronger model are retained as valid hard examples. Overall, this filtering pipeline removes roughly $70\%$--$90\%$ of raw candidates before trajectory synthesis, substantially improving both verifiability and training utility. We employ Qwen-Max in an LLM-as-Judge framework to evaluate the correctness of tasks.

\section{Details of the Plan-then-Execute Pipeline}
\label{app:plan_execute}

To synthesize executable trajectories for both single-tool and multi-tool supervision, we adopt a plan-then-execute pipeline that separates high-level tool planning from grounded environment interaction. This decomposition improves controllability during synthesis while ensuring that the final trajectory is grounded in actual tool-returned observations.

\paragraph{Planner.}
We use a strong thinking VLM, Qwen3-VL-235B-A22B-Thinking, as the planner. Given the user query, visual input, and ground-truth answer, the planner reconstructs a plausible sequence of tool calls that could lead to the target answer. Importantly, the generated tool arguments at this stage are used only as regularizing hints for plan structure; they do not directly determine the final executed trajectory. Actual execution is delegated to a separate executor model that interacts with the tool environment step by step. The ground-truth answer thus only guides the planner and filters out answer-inconsistent trajectories during synthesis; the executor and the final trained model operate without it, and it is not available at inference. The exact planning prompt is shown in Prompt~\ref{prompt:tool_panning}.

\begin{prompt*}[t]
\centering
\begin{tcolorbox}[
 width=\textwidth,
 colback=black!3!white,
 colframe=black!75!white,
 rounded corners,
 arc=1mm,
 boxrule=0.2mm,
 title=Prompt for tool planning
]
\footnotesize
\textbf{Role}

You are a tool-orchestration assistant. Your task is to generate a sequence of tool calls that would logically lead to the provided ground-truth answer based on the user query.

\textbf{Inputs}

\textbf{User Query:} The question or task requested by the user. \\
\textbf{Ground Truth:} The correct final information or result that must be reached.

\textbf{Task}

Reverse-engineer the necessary steps required to arrive at the ground truth. For each step, select the most appropriate tool and provide the exact parameters.

\textbf{Output Format (Strict)}

Return only a single JSON array of objects. Each object must contain:
\begin{itemize}
 \item \texttt{"description"}: A concise explanation of why this step is taken to reach the ground truth.
 \item \texttt{"tool\_name"}: The exact name of the tool.
 \item \texttt{"parameters"}: A dictionary of arguments for the tool. Ensure that types such as \texttt{int}, \texttt{float}, \texttt{list}, and \texttt{str} are correct.
\end{itemize}

\textbf{Allowed Tools}

\texttt{image\_zoom\_in\_tool}: Zoom in on a specific region of an image by cropping it based on a bounding box (\texttt{bbox}) and an optional object label. \\
Parameters: \texttt{\{"bbox\_2d": [x1, y1, x2, y2], "label": "string", "img\_idx": int\}}

\texttt{image\_rotate\_tool}: Rotate an image by a specified angle. If required, you may zoom into specific regions and apply rotation. Positive values rotate counter-clockwise; negative values rotate clockwise. \\
Parameters: \texttt{\{"angle": float, "img\_idx": int, "bbox\_2d": [x1, y1, x2, y2]\}}

\texttt{image\_enhance\_tool}: Enhance an image by adjusting contrast (\texttt{factor}) and/or brightness (\texttt{gamma}). Contrast $> 1.0$ increases detail visibility; gamma $< 1.0$ brightens the image. \\
Parameters: \texttt{\{"img\_idx": int, "contrast": float, "gamma": float\}}

\texttt{image\_code\_tool}: Generate and execute Python code. If \texttt{img\_idx} is provided, the image is available as variable \texttt{img} in the code scope. This tool may be used for geometry drawing, computation for math/chart/document tasks, but not OCR. \\
Parameters: \texttt{\{"code": "string", "img\_idx": int\}}

\texttt{image\_edit\_tool}: Edit an existing source image based on textual instructions. \\
Parameters: \texttt{\{"img\_idx": int, "prompt": "string", "negative\_prompt": "string"\}}

\texttt{multimodal\_search}: A unified search tool. Provide \texttt{query} for standard web search, or \texttt{img\_idx} for reverse image search. \\
Parameters: \texttt{\{"query": "string", "img\_idx": int\}}

\texttt{image\_mark\_tool}: Mark an image using points \texttt{[[x,y]]}, boxes \texttt{[[x1,y1,x2,y2]]}, or negative marks for exclusion. This tool can be used for counting or highlighting. \\
Parameters: \texttt{\{"img\_idx": int, "points": list, "boxes": list, "negative\_marks": list, "label": "string"\}}

\texttt{none}: Use this for final reasoning or when no external tool is required. \\
Parameters: \texttt{\{\}}

\textbf{Rules}

Do not include any preamble or explanation. Do not output phrases such as ``Here is the plan,'' and do not use Markdown code fences such as \texttt{```json}. Output only the raw JSON string, starting with \texttt{[} and ending with \texttt{]}.

The generated steps must be consistent with the ground truth. If the ground truth contains a specific detail that would require zooming, searching, enhancement, or another operation, the corresponding tool call must appear in the sequence.

All \texttt{img\_idx} values start at 0. Subsequent tool outputs increment the index. Ensure that the JSON is valid and parsable.

\textbf{Example Output Structure}

\texttt{[} \\
\texttt{\{"description": "Reason for step 1", "tool\_name": "multimodal\_search", "parameters": \{"query": "..."\}\},} \\
\texttt{\{"description": "Reason for step 2", "tool\_name": "none", "parameters": \{\}\}} \\
\texttt{]}

\textbf{Your Task}

\texttt{[Input]: <image>} \\
\texttt{[User Query]: \{question\}} \\
\texttt{[Ground Truth]: \{ground\_truth\}}

Based on the rules above, analyze the input and provide the output.
\end{tcolorbox}
\caption{Prompt for tool planning}
\label{prompt:tool_panning}
\end{prompt*}

\paragraph{Executor.}
We use Qwen3-VL-235B-A22B-Instruct as the executor. Conditioned on the current interaction history and the planner-provided tool sequence, the executor predicts grounded reasoning and executable tool arguments step by step. Because execution occurs in the actual tool environment, later actions depend on returned observations rather than on hallucinated intermediate states. This is important for synthesizing trajectories that reflect realistic multi-turn visual reasoning.

\begin{table*}[t]
\centering
\small
\caption{Statistics of the SFT trajectory bank by data type.
We report the number of trajectories, the average number of tool calls,
the average number of messages per trajectory, and the average trajectory length in characters.}
\label{tab:sft_trajectory_stats}
\begin{tabular}{lrrrr}
\toprule
Trajectory Type & \# Trajectories & Avg. Tool Calls & Avg. Messages & Avg. Length (Chars) \\
\midrule
Single-Tool & 28,590 & 1.0602 & 5.1220 & 6,627.00 \\
Multi-Tool & 4,945 & 2.6849 & 8.3699 & 5,795.10 \\
Diverse & 6,437 & 2.8614 & 8.7229 & 5,672.28 \\
\bottomrule
\end{tabular}
\end{table*}

\section{Details of Diverse Tool Contexts}
\label{app:dtc}

To improve generalization beyond a fixed tool interface, we additionally synthesize diverse tool-context trajectories by reinstantiating the same underlying visual operation under alternative schemas. Concretely, given an initial image state, a transformed result image, and a textual description of the intended operation, we ask a strong VLM to either match the transformation to an existing tool from a candidate pool or define a new tool schema if no suitable tool exists.

We use Qwen3-VL-235B-A22B-Instruct for this schema reinstantiation process. The prompt is shown as Prompt~\ref{prompt:diverse_tool_contexts}. This procedure encourages the model to associate a tool with its \emph{functional affordance} rather than memorizing a fixed tool name or argument schema. In particular, by preserving the underlying visual transition while varying schema surface forms, we obtain supervision that is better aligned with zero-shot transfer to unseen tool interfaces.

\noindent\textbf{Executable and reinstantiated tools.} Our pipeline involves tools of three kinds. \emph{Base training tools} (e.g., zoom, rotate, enhance, code, mark, search) are backed by executors and are actually run during plan-then-execute synthesis, producing genuine observations $o_k$. \emph{Reinstantiated tools} (e.g., the schemas in Table~\ref{tab:novel_tool_examples}) are not backed by new executors; they relabel an already-observed transition at the schema level, keeping the observation $o_k$ inherited from a base-tool execution or from real image pairs $(o_{k-1}, o_k)$ in implicit-reasoning datasets, so that only the tool's surface description and arguments vary. \emph{Inference-time tools} (Table~\ref{tab:unseen_tool_inventory}, e.g., flip, brightness, histeq) are again executable, but are provided to the model only through natural-language schema descriptions at test time.

\begin{prompt*}[t]
\centering
\begin{tcolorbox}[
 width=\textwidth,
 colback=black!3!white,
 colframe=black!75!white,
 rounded corners,
 arc=1mm,
 boxrule=0.2mm,
 title=Prompt for diverse tool contexts reinstantiation
]
\footnotesize
\textbf{Role}

You are an expert AI assistant specializing in visual analysis and tool generation. Your primary goal is to analyze an image transformation, consult a list of available tools, and then generate the necessary tool definition (if a new tool is required) and the corresponding tool call to achieve the result.

\textbf{Task Instructions}

Follow these steps:

\textbf{1. Analyze the inputs.} Carefully examine the Initial Image, the Result Image, and the user's Thought Process to understand the exact operation performed.

\textbf{2. Consult the tool pool.} Check the Tool Candidate Pool. If a tool in the pool perfectly matches the required operation, you must use it. Do not redefine an existing tool.

\textbf{3. Decide and define.}
If a suitable tool already exists, directly generate the corresponding tool call. \\
If no suitable tool exists, first define a new logical tool that can perform the transformation. The tool definition must follow standard JSON format and contain a name, a description, and a list of parameters with their types and descriptions.

\textbf{4. Generate the output.} Your final output must be a single JSON object containing exactly two keys: \texttt{tool\_definition} and \texttt{tool\_call}.

\textbf{Output Requirements}

\texttt{tool\_definition}: A JSON object defining the new tool. If you are using an existing tool from the pool, this value must be an empty string. \\
\texttt{tool\_call}: A JSON object representing the function call, with appropriate arguments to transform the Initial Image into the Result Image.

\textbf{Important Constraint}

Do not include any exact coordinates in the parameters.

\textbf{Your Turn}

\textbf{[Inputs]} \\
\texttt{Initial Image: <image>} \\
\texttt{Result Image: <image>} \\
\texttt{Thought Process: \{\}} \\
\texttt{Tool Candidate Pool: \{\}}

\textbf{[Output]}
\end{tcolorbox}
\caption{Prompt for diverse tool contexts reinstantiation}
\label{prompt:diverse_tool_contexts}
\end{prompt*}

\paragraph{Representative synthesized schemas.}
To illustrate the diversity induced by our tool-context reinstantiation procedure, Table~\ref{tab:novel_tool_examples} summarizes three representative synthesized tool schemas. They vary substantially in abstraction level and frequency: some correspond to recurring latent operations that appear across many synthesized trajectories, while others are rare and highly specialized. As shown, the generated tools range from broad scene-editing operations, to visually grounded identification tools, to narrowly scoped mathematical operators. This diversity is precisely what we aim to expose during training: the model should infer tool affordances from schema semantics rather than memorize a small fixed set of canonical tool names.

\begin{table*}[t]
\centering
\caption{Representative synthesized tool schemas from diverse tool-context generation.}
\label{tab:novel_tool_examples}
\small
\renewcommand{\arraystretch}{1.12}
\setlength{\tabcolsep}{4pt}
\begin{tabular}{p{4.5cm} p{4.4cm} p{5.6cm}}
\toprule
\textbf{Tool Name} & \textbf{Description} & \textbf{Key Parameters} \\
\midrule
\texttt{add\_objects} &
Adds new 3D objects to the scene. &
\texttt{image\_idx}, \texttt{objects} (\texttt{shape}, \texttt{color}, \texttt{count}) \\[6pt]

\texttt{identify\_and\_} \texttt{highlight\_objects} &
Identifies described objects and highlights them with colored boxes. &
\texttt{image\_idx}, \texttt{object\_description}, \texttt{num\_objects} \\[6pt]

\texttt{calculate\_triangle\_side} &
Calculates a target side in a 30-60-90 triangle from a known side. &
\texttt{image\_idx}, \texttt{given\_side\_length}, \texttt{given\_side\_type}, \texttt{target\_side\_type} \\
\bottomrule
\end{tabular}
\end{table*}

The frequent schema \texttt{add\_objects} reflects a broad latent operation that recurs across many implicit visual transformation instances. In contrast, \texttt{identify\_and\_highlight\_objects} is a visually grounded localization tool, while \texttt{calculate\_triangle\_side} represents a rare mathematical reasoning operator. Together, these examples show that our synthesized schema space is not limited to minor renamings of existing tools, but spans a broad range of functional abstractions. The full JSON schemas of several representative synthesized tools are listed below.

\begin{prompt*}[t]
\centering
\begin{tcolorbox}[
 colback=black!3!white,
 colframe=black!75!white,
 rounded corners,
 arc=1mm,
 boxrule=0.2mm,
 title=Representative Novel Tool Schema: \texttt{add\_objects}
]
\scriptsize\ttfamily
\{\\
\hspace*{1em}"name": "add\_objects",\\
\hspace*{1em}"description": "Adds new 3D objects to the scene.",\\
\hspace*{1em}"parameters": \{\\
\hspace*{2em}"type": "object",\\
\hspace*{2em}"properties": \{\\
\hspace*{3em}"image\_idx": \{\\
\hspace*{4em}"type": "number",\\
\hspace*{4em}"description": "The index of the image to add objects to\\
\hspace*{4em}(starting from 0)."\\
\hspace*{3em}\},\\
\hspace*{3em}"objects": \{\\
\hspace*{4em}"type": "array",\\
\hspace*{4em}"items": \{\\
\hspace*{5em}"type": "object",\\
\hspace*{5em}"properties": \{\\
\hspace*{6em}"shape": \{\\
\hspace*{7em}"type": "string",\\
\hspace*{7em}"description": "The geometric shape of the object\\
\hspace*{7em}(e.g., `pyramid', `sphere', `cone')."\\
\hspace*{6em}\},\\
\hspace*{6em}"color": \{\\
\hspace*{7em}"type": "string",\\
\hspace*{7em}"description": "The color of the object\\
\hspace*{7em}(e.g., `orange', `red', `blue')."\\
\hspace*{6em}\},\\
\hspace*{6em}"count": \{\\
\hspace*{7em}"type": "number",\\
\hspace*{7em}"description": "The number of objects of this shape and\\
\hspace*{7em}color to add."\\
\hspace*{6em}\}\\
\hspace*{5em}\},\\
\hspace*{5em}"required": ["shape", "color", "count"]\\
\hspace*{4em}\},\\
\hspace*{4em}"description": "A list of objects to add, each defined by\\
\hspace*{4em}its shape, color, and count."\\
\hspace*{3em}\}\\
\hspace*{2em}\},\\
\hspace*{2em}"required": ["image\_idx", "objects"]\\
\hspace*{1em}\}\\
\}
\end{tcolorbox}
\label{fig:add_objects_schema}
\end{prompt*}

\begin{prompt*}[t]
\centering
\begin{tcolorbox}[
 colback=black!3!white,
 colframe=black!75!white,
 rounded corners,
 arc=1mm,
 boxrule=0.2mm,
 title=Representative Novel Tool Schema: \texttt{identify\_and\_highlight\_objects}
]
\scriptsize\ttfamily
\{\\
\hspace*{1em}"name": "identify\_and\_highlight\_objects",\\
\hspace*{1em}"description": "Identifies specific objects in an image\\
\hspace*{1em}based on a description and highlights them with\\
\hspace*{1em}colored bounding boxes.",\\
\hspace*{1em}"parameters": \{\\
\hspace*{2em}"type": "object",\\
\hspace*{2em}"properties": \{\\
\hspace*{3em}"image\_idx": \{\\
\hspace*{4em}"type": "number",\\
\hspace*{4em}"description": "The index of the image to process\\
\hspace*{4em}(starting from 0)."\\
\hspace*{3em}\},\\
\hspace*{3em}"object\_description": \{\\
\hspace*{4em}"type": "string",\\
\hspace*{4em}"description": "A clear description of the objects to be\\
\hspace*{4em}identified and highlighted."\\
\hspace*{3em}\},\\
\hspace*{3em}"num\_objects": \{\\
\hspace*{4em}"type": "number",\\
\hspace*{4em}"description": "The number of objects to identify and\\
\hspace*{4em}highlight."\\
\hspace*{3em}\}\\
\hspace*{2em}\},\\
\hspace*{2em}"required": ["image\_idx", "object\_description",\\
\hspace*{2em}"num\_objects"]\\
\hspace*{1em}\}\\
\}
\end{tcolorbox}
\label{fig:highlight_objects_schema}
\end{prompt*}

\begin{prompt*}[t]
\centering
\begin{tcolorbox}[
 colback=black!3!white,
 colframe=black!75!white,
 rounded corners,
 arc=1mm,
 boxrule=0.2mm,
 title=Representative Novel Tool Schema: \texttt{calculate\_triangle\_side}
]
\scriptsize\ttfamily
\{\\
\hspace*{1em}"name": "calculate\_triangle\_side",\\
\hspace*{1em}"description": "Calculates the length of a specific side in a\\
\hspace*{1em}30-60-90 triangle given the length of another side\\
\hspace*{1em}and the type of side it is (shorter leg, longer leg,\\
\hspace*{1em}or hypotenuse).",\\
\hspace*{1em}"parameters": \{\\
\hspace*{2em}"type": "object",\\
\hspace*{2em}"properties": \{\\
\hspace*{3em}"image\_idx": \{\\
\hspace*{4em}"type": "number",\\
\hspace*{4em}"description": "The index of the image containing the\\
\hspace*{4em}triangle (starting from 0)."\\
\hspace*{3em}\},\\
\hspace*{3em}"given\_side\_length": \{\\
\hspace*{4em}"type": "number",\\
\hspace*{4em}"description": "The length of the known side."\\
\hspace*{3em}\},\\
\hspace*{3em}"given\_side\_type": \{\\
\hspace*{4em}"type": "string",\\
\hspace*{4em}"description": "The type of the known side: `shorter\_leg',\\
\hspace*{4em}`longer\_leg', or `hypotenuse'."\\
\hspace*{3em}\},\\
\hspace*{3em}"target\_side\_type": \{\\
\hspace*{4em}"type": "string",\\
\hspace*{4em}"description": "The type of the side to be calculated:\\
\hspace*{4em}`shorter\_leg', `longer\_leg', or `hypotenuse'."\\
\hspace*{3em}\}\\
\hspace*{2em}\},\\
\hspace*{2em}"required": ["image\_idx", "given\_side\_length",\\
\hspace*{2em}"given\_side\_type", "target\_side\_type"]\\
\hspace*{1em}\}\\
\}
\end{tcolorbox}
\label{fig:triangle_side_schema}
\end{prompt*}

\begin{prompt*}[t]
\centering
\begin{tcolorbox}[
 colback=black!3!white,
 colframe=black!75!white,
 rounded corners,
 arc=1mm,
 boxrule=0.2mm,
 title=Representative Novel Tool Schema: \texttt{image\_translate\_tool}
]
\scriptsize\ttfamily
\{\\
\hspace*{1em}"name": "image\_translate\_region",\\
\hspace*{1em}"description": "Find regions containing texts, and translate the\\
\hspace*{1em}text within the region. Overlay the translated text\\
\hspace*{1em}back onto the original position. It automatically\\
\hspace*{1em}removes the original text and matches the background.",\\
\hspace*{1em}"parameters": \{\\
\hspace*{2em}"type": "object",\\
\hspace*{2em}"properties": \{\\
\hspace*{3em}"img\_idx": \{\\
\hspace*{4em}"type": "number",\\
\hspace*{4em}"description": "The index of the image containing foreign\\
\hspace*{4em}text (starting from 0)."\\
\hspace*{3em}\},\\
\hspace*{3em}"translation\_tasks": \{\\
\hspace*{4em}"type": "array",\\
\hspace*{4em}"description": "A list of translation tasks, each pairing a\\
\hspace*{4em}specific region with its translated text.",\\
\hspace*{4em}"items": \{\\
\hspace*{5em}"type": "object",\\
\hspace*{5em}"properties": \{\\
\hspace*{6em}"bbox": \{\\
\hspace*{7em}"type": "array",\\
\hspace*{7em}"description": "The bounding box [x1, y1, x2, y2] that fully\\
\hspace*{7em}cover the translated text region.",\\
\hspace*{7em}"items": \{"type": "number"\},\\
\hspace*{7em}"minItems": 4,\\
\hspace*{7em}"maxItems": 4\\
\hspace*{6em}\},\\
\hspace*{6em}"text": \{\\
\hspace*{7em}"type": "string",\\
\hspace*{7em}"description": "The translated text to be placed in this\\
\hspace*{7em}specific bounding box."\\
\hspace*{6em}\}\\
\hspace*{5em}\},\\
\hspace*{5em}"required": ["bbox", "text"]\\
\hspace*{4em}\}\\
\hspace*{3em}\}\\
\hspace*{2em}\},\\
\hspace*{2em}"required": ["img\_idx", "translation\_tasks"]\\
\hspace*{1em}\}\\
\}
\end{tcolorbox}
\label{fig:translate_text_region_schema}
\end{prompt*}

\section{Trajectory Filtering and Augmentation}
\label{app:traj_filtering}

To improve the quality of synthesized supervision, we apply several post-processing rules to generated trajectories before adding them to the final SFT training bank.

\paragraph{Answer consistency.}
We discard trajectories whose final answer does not match the ground-truth annotation. This ensures that the synthesized reasoning path remains aligned with the original instance label.

\paragraph{Structural validity.}
We remove trajectories if any tool call is malformed, violates the corresponding tool schema, or fails during execution. This step is essential because invalid or non-executable calls can introduce spurious supervision for argument prediction and multi-turn state tracking.

\paragraph{Efficiency constraint for single-tool trajectories.}
For single-tool trajectories, we further filter out cases containing redundant repeated calls to the same tool. This encourages concise and precise invocation behavior rather than unnecessarily iterative patterns that are unlikely to be useful as foundational supervision.

\paragraph{Tool-pool augmentation.}
To expose the model to richer and more realistic tool environments, we randomly augment the candidate tool pool of each trajectory with additional distractor tools. This prevents the model from overfitting to minimal task-specific tool sets and encourages more robust tool selection under larger candidate spaces.

\paragraph{Trajectory Statistics.}
Table~\ref{tab:sft_trajectory_stats} summarizes the composition of the SFT trajectory bank.
Single-tool trajectories constitute the majority of the supervision data, providing stable grounding for visual tool invocation.
Multi-tool and diverse-context trajectories are fewer in number but involve substantially longer interaction histories and more tool calls on average,
reflecting their higher compositional and adaptive complexity.

\section{Tool Reward}
\label{app:process_reward}

In the RL stage, we supplement the standard answer-correctness reward with a tool reward designed to capture whether the trajectory uses tools in a grounded and adaptive manner. The reward is computed by a lightweight critic that evaluates the entire rollout trajectory along five binary dimensions. The exact evaluation prompt is given in Prompt~\ref{prompt:process_aware_prompt}. We leverage Qwen3.5-Plus as the LLM-as-Judge to evaluate the reasoning process.

\begin{prompt*}[t]
\centering
\begin{tcolorbox}[
    width=\textwidth,
    colback=black!3!white,
    colframe=black!75!white,
    rounded corners,
    arc=1mm,
    boxrule=0.2mm,
    title=Prompt for Tool Reward
]
\footnotesize
\textbf{Role and Goal}

You are an expert evaluator. Your task is to perform a binary (0 or 1) assessment of an AI assistant's tool-use behavior across a multi-turn trajectory. You must judge the trajectory based on five specific dimensions, where 1 means \textit{Pass/Yes} and 0 means \textit{Fail/No}.

\textbf{Evaluation Dimensions (0--1 Scoring)}

\textbf{Plan Coherence.}
1 (Pass): The assistant's thought process follows a clear, step-by-step logical chain. Each step or tool call is a reasonable derivation from the previous information. \\
0 (Fail): There are logical leaps, non-sequiturs, or the reasoning contradicts the provided information or tool outputs.

\textbf{Call Efficiency.}
1 (Pass): Every tool call is necessary and purposeful. The assistant avoids repeating the same call with the same parameters and does not get stuck in thinking loops. \\
0 (Fail): There are redundant tool calls, or the assistant repeats the same reasoning/actions without making progress.

\textbf{Feedback Responsiveness.}
1 (Pass): The assistant demonstrates self-correction or re-planning. For example, it uses phrases such as ``Wait,'' ``Actually,'' or ``No, that's incorrect, I should try...'' when tool output is unexpected or an error occurs. \\
0 (Fail): The assistant blindly follows a flawed plan or ignores errors/contradictions in the feedback, showing no signs of internal reflection.

\textbf{Observation Fidelity.}
1 (Pass): The assistant accurately extracts and uses data from tool outputs or images. It does not hallucinate numbers, dates, or facts that are not present in the feedback. \\
0 (Fail): The assistant misreads tool outputs, hallucinates information not present in the feedback, or incorrectly interprets error messages.

\textbf{Visual Grounding.}
1 (Pass): The assistant's tool parameters are strictly consistent with the visual content of the image. For example, if it zooms into or marks a region, the coordinates match the intended visual target. \\
0 (Fail): Tool arguments reference elements not present in the image, target wrong regions, or make claims about the image that are visually demonstrably false.

\textbf{Output Format}

You must provide your evaluation in a single, clean JSON object. Do not include any text before or after the JSON block.

\texttt{\{}\\
\texttt{\ \ "plan\_coherence": <0 or 1>,}\\
\texttt{\ \ "call\_efficiency": <0 or 1>,}\\
\texttt{\ \ "feedback\_responsiveness": <0 or 1>,}\\
\texttt{\ \ "observation\_fidelity": <0 or 1>,}\\
\texttt{\ \ "visual\_grounding": <0 or 1>}\\
\texttt{\}}

\textbf{Your Task}

\texttt{\{\}}

\textbf{Output}
\end{tcolorbox}
\caption{Prompt for tool reward.}
\label{prompt:process_aware_prompt}
\end{prompt*}

Given a rollout trajectory $\tau$, the critic outputs five binary scores, one for each criterion above. We define the tool reward as their average:
$$
R_{\mathrm{tool}}(\tau)=\frac{1}{5}\sum_{m=1}^{5} c_m(\tau),
$$
where each $c_m(\tau)\in\{0,1\}$. This reward does not directly incentivize making more tool calls; instead, it rewards whether the trajectory uses tool outputs coherently, efficiently, and faithfully in subsequent reasoning.

\paragraph{Role of each dimension.}
The five binary criteria target complementary aspects of tool-use behavior that the reward signal sharpens during RL. \emph{Plan Coherence} checks that each tool call follows logically from the preceding interaction history, discouraging arbitrary jumps or contradictions between the model's plan and the tool environment's feedback. \emph{Call Efficiency} penalizes redundant or repeated calls with the same parameters, directly targeting E3-type (redundant/unnecessary call) behavior. \emph{Feedback Responsiveness} rewards the model for adjusting its next action when a tool output is unexpected or an execution error occurs, rather than blindly continuing a flawed trajectory. This addresses the E4 pattern where a tool is invoked but its result is ignored. \emph{Observation Fidelity} checks that the model accurately uses data from tool outputs without hallucinating numbers or facts not present in the returned observation, targeting E5-type failures where valid tool results are misinterpreted. Finally, \emph{Visual Grounding} verifies that tool arguments are grounded in the actual image content (e.g., zoom coordinates match the intended region), addressing E2 errors where parameters are syntactically valid but visually mislocated. Together, these five dimensions provide a holistic shaping signal that sharpens tool-use behavior from low-level call validity (E1) through execution discipline (E3, E4) to high-level visual grounding (E2, E5).

\section{RL Training Data}
\label{app:rl_training_data}

For RL, we collect training instances from existing multimodal reasoning and tool-use data sources, including ChartVerse~\citep{liu2026chartverse}, DeepEyes~\citep{zheng2025deepeyes}, DRIM~\citep{yang2025deep}, and VisualProbe~\citep{lai2025mini}. Compared with the SFT trajectory bank, the RL data is intended to support policy refinement under environment interaction rather than purely supervised imitation.

\begin{figure}[t!]
 \centering
 \includegraphics[width=0.9\columnwidth]{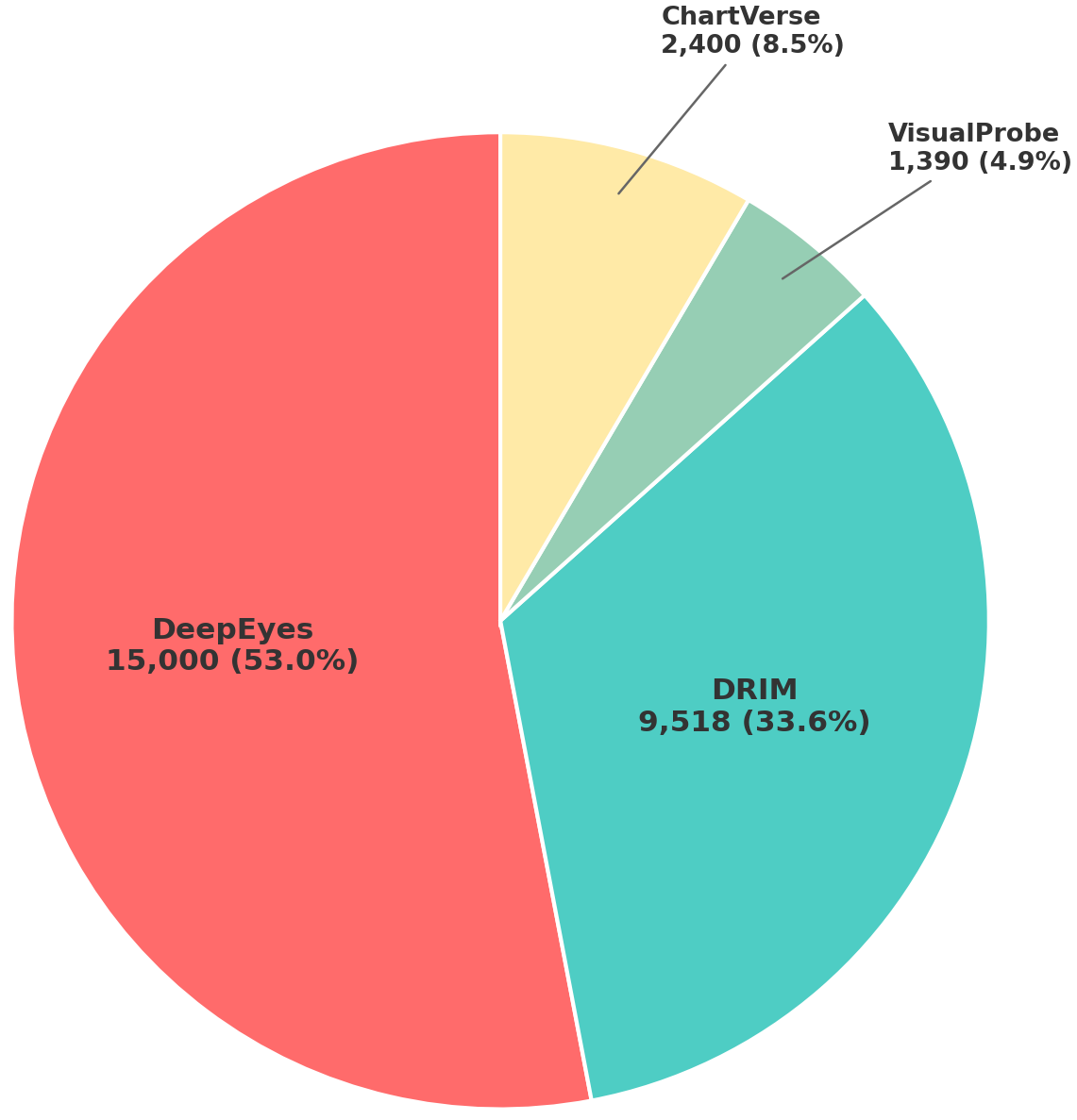}
 \caption{\textbf{Composition of the RL training data.} The figure summarizes the distribution of RL training instances across tool-use categories.}
 \label{fig:rl_data}
 \vspace{-0.6cm}
\end{figure}

To improve training stability and avoid spending optimization budget on excessively noisy or unsalvageable instances, we further remove samples that the base policy model, i.e. Qwen3-VL-8B-Instruct, consistently fails to solve in 8 attempts. We employ Qwen-Max in an LLM-as-Judge framework to evaluate the correctness of task completion. This filtering step yields a moderate-difficulty RL set that is better suited for learning adaptive tool-use behaviors. The resulting RL dataset contains 28K training samples. Its composition is illustrated in Fig.~\ref{fig:rl_data}. We will release the full dataset composition, source breakdown, and filtering scripts upon acceptance.

\section{RL Training Dynamics}
\label{sec:rl_dynamics}

This section reports the RL training dynamics that were referenced in the main paper but omitted from the main text due to space constraints. All settings and notation follow those of the main paper.

\begin{figure*}[t!]
 \centering
 \includegraphics[width=\textwidth]{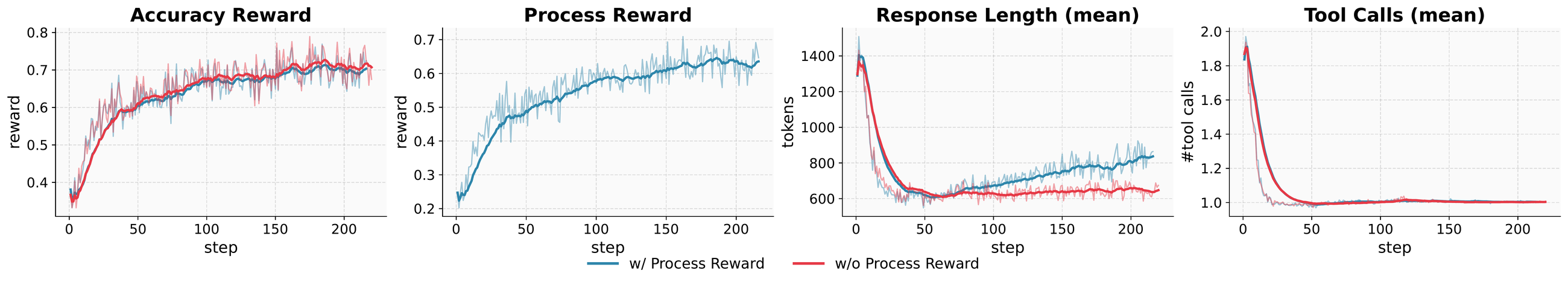}
 \caption{\textbf{Training dynamics of RL with and without tool reward.} Accuracy reward, tool reward, average response length, and average number of tool calls per training step.}
 \label{fig:rl_dynamics}
\end{figure*}

Figure~\ref{fig:rl_dynamics} compares the RL training dynamics with and without the tool reward. Although the two variants exhibit similar trends in accuracy reward, the model trained with the tool reward maintains a consistently higher tool reward throughout training, and its responses become longer in the middle and later stages of RL. This is consistent with the picture in the main paper: the tool reward does not change the outcome signal itself, but shapes the policy toward trajectories that make more substantive use of tool feedback.

\section{Inference-Time Unseen Tools}
\label{app:unseen_tools}

To evaluate schema-level generalization beyond the tool interfaces observed during training, we additionally introduce a pool of previously unseen tools at inference time. These tools cover a range of image transformation and image-processing operations, including geometric manipulation, color editing, enhancement, segmentation, and restoration. Table~\ref{tab:unseen_tool_inventory} lists the unseen tools used in our evaluation setup.

\begin{table}[t]
\centering
\caption{Previously unseen tools introduced at inference time.}
\label{tab:unseen_tool_inventory}
\small
\renewcommand{\arraystretch}{1.12}
\setlength{\tabcolsep}{4pt}
\begin{tabular}{p{0.42\columnwidth} p{0.48\columnwidth}}
\toprule
\textbf{Tool Name} & \textbf{Function} \\
\midrule
\texttt{flip} & Flip an image horizontally or vertically \\
\texttt{color} & Modify or adjust image colors \\
\texttt{brightness} & Adjust image brightness \\
\texttt{histeq} & Apply histogram equalization \\
\texttt{binarize} & Convert an image into a binary image \\
\texttt{floodfill} & Fill a connected image region from a seed point \\
\texttt{denoise} & Remove image noise \\
\texttt{sharpen} & Sharpen image details \\
\texttt{inpaint} & Restore or fill missing image regions \\
\texttt{inrange\_color} & Segment pixels within a specified color range \\
\texttt{morph} & Apply morphological operations such as dilation or erosion \\
\bottomrule
\end{tabular}
\end{table}

These tools are not included in the training-time tool pool. Instead, they are only provided through natural-language schema descriptions at inference time, making them a direct test of whether the model can infer tool affordances from previously unseen interfaces.

\section{Tool Call Statistics}
\label{app:tool_stats}

To better characterize the learned tool-use policy after RL, we report aggregate tool-call statistics for both tools seen during training and tools unseen during training. Here, ``Success Rate'' refers to schema-level execution success: the fraction of invocations for which the model produces a schema-valid call, a correct tool name with well-formed arguments, that the environment executes without error.

\begin{table}[t!]
\centering
\caption{RL tool call statistics for tools seen during training.}
\label{tab:tool_seen}
\begin{tabular}{lcc}
\toprule
\textbf{Tool} & \textbf{Avg. Call} & \textbf{Success} \\
 & \textbf{Prop. (\%)} & \textbf{Rate (\%)} \\
\midrule
zoom\_in & 78.9 & 99.7 \\
code & 10.5 & 63.4 \\
enhance & 3.2 & 99.7 \\
rotate & 0.9 & 100.0 \\
mark & 0.8 & 95.4 \\
edit & 0.1 & 88.9 \\
\midrule
no\_tool & 8.1 & - \\
\bottomrule
\end{tabular}
\end{table}

\begin{table}[t!]
\centering
\caption{RL tool call statistics for tools unseen during training.}
\label{tab:tool_unseen}
\begin{tabular}{lcc}
\toprule
\textbf{Tool} & \textbf{Avg. Call} & \textbf{Success} \\
 & \textbf{Prop. (\%)} & \textbf{Rate (\%)} \\
\midrule
brightness & 1.7 & 73.9 \\
inrange\_color & 1.7 & 84.5 \\
morph & 0.6 & 98.3 \\
sharpen & 0.5 & 100.0 \\
histeq & 0.3 & 93.9 \\
\midrule
others & 0.7 & 97.1 \\
\bottomrule
\end{tabular}
\end{table}

These statistics show that the learned policy remains dominated by a small number of highly reliable visual tools, especially zoom-based inspection. More importantly, the model also demonstrates strong transfer to previously unseen tool schemas. Although unseen tools are invoked less frequently overall, their success rates are consistently high once selected: several unseen tools, including \texttt{inrange\_color}, \texttt{morph}, \texttt{sharpen}, and \texttt{histeq}, achieve success rates between $84.5\%$ and $100.0\%$. Even the more challenging \texttt{brightness} tool reaches a success rate of $73.9\%$. These results suggest that the model is able to infer tool affordances from schema descriptions and integrate them into ongoing reasoning, rather than relying solely on memorized training-time APIs. This supports our claim that VC-Tooler learns visual tool use as a compositional and adaptive capability with meaningful zero-shot transfer to novel tool interfaces.

\begin{figure*}[t]
    \centering
    \begin{subfigure}[b]{0.47\textwidth}
        \centering
        \includegraphics[width=\linewidth]{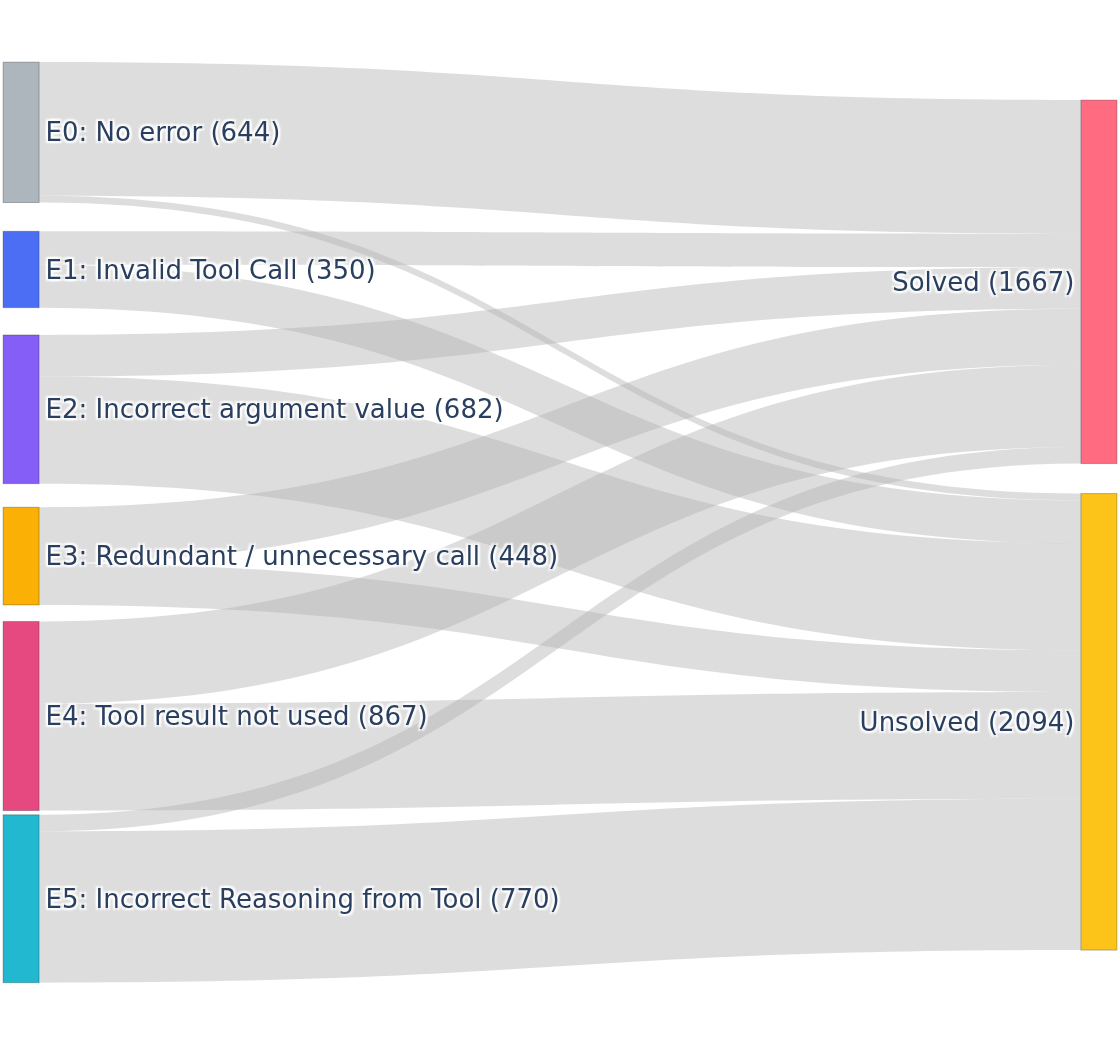}
        \caption{After SFT}
        \label{fig:error_sft}
    \end{subfigure}
    \hfill
    \begin{subfigure}[b]{0.45\textwidth}
        \centering
        \includegraphics[width=\linewidth]{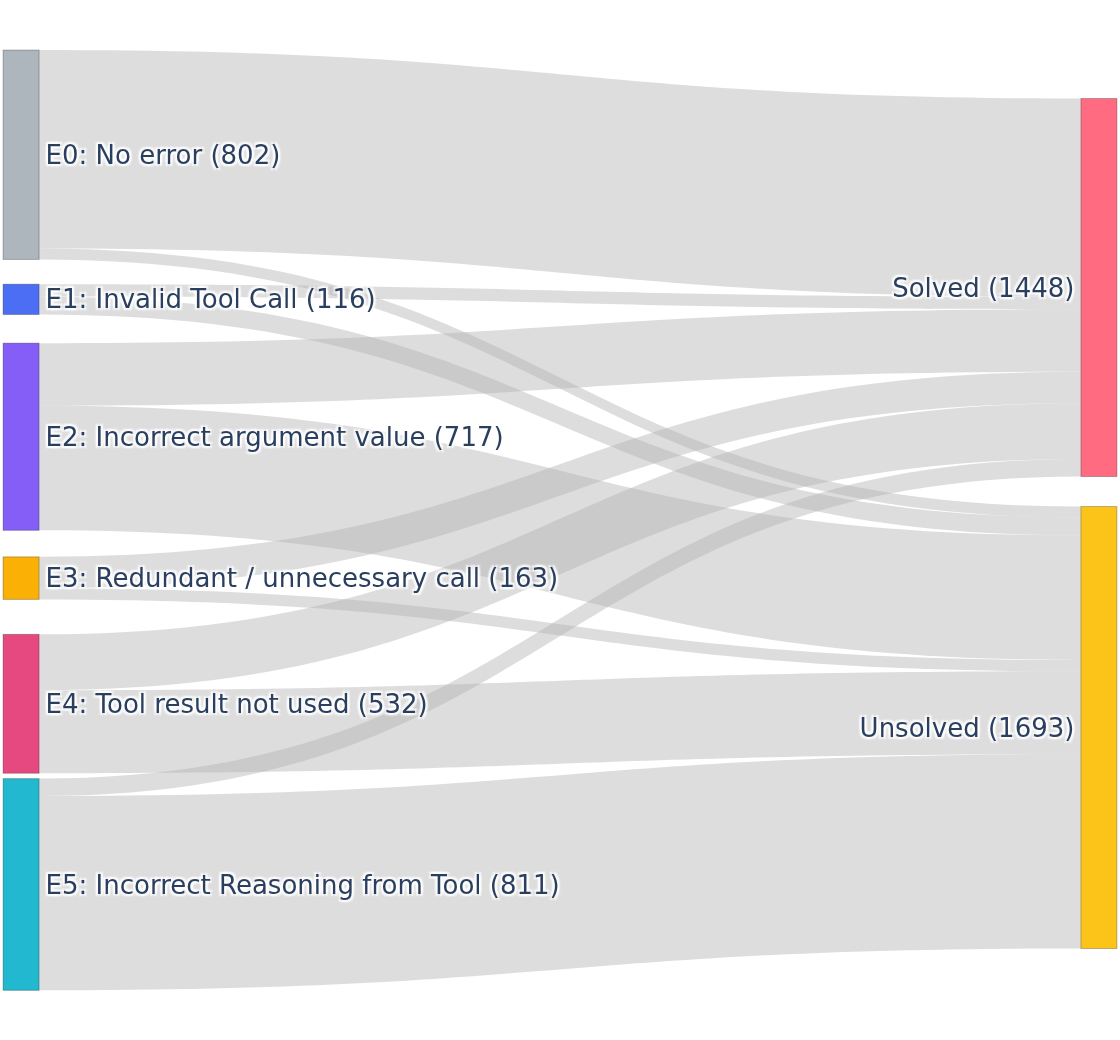}
        \caption{After RL}
        \label{fig:error_rl}
    \end{subfigure}
    \caption{\textbf{Error analysis before and after RL.} Sankey diagrams linking six error categories (E0--E5) to the solved/unsolved outcome, (a) after cold-start SFT and (b) after RL. Numbers in parentheses give the count of trajectories in each category.}
    \label{fig:error_analysis}
\end{figure*}

\section{Error Analysis}
\label{app:error_analysis}

To understand where the remaining failures come from, we categorize each rollout by its dominant error type and trace how the six categories flow to the final solved/unsolved outcome. We use six labels: \emph{E0}~no tool-use error; \emph{E1}~invalid tool call, covering an illegal invocation structure (malformed JSON, missing fields), an argument name absent from the tool schema, or an argument value with an invalid type or format (e.g., non-numeric or out-of-range coordinates); \emph{E2}~incorrect argument value, where the call is syntactically valid but semantically wrong (e.g., zooming into the wrong region); \emph{E3}~redundant or unnecessary call; \emph{E4}~tool result not used, where the tool is invoked and returns a valid result but the model ignores it in subsequent reasoning; and \emph{E5}~incorrect reasoning from a tool output, where the tool returns a valid result but the model reasons or answers incorrectly from it. Figure~\ref{fig:error_analysis} compares the error distributions after cold-start SFT and after RL.

Comparing the two stages reveals what RL actually fixes. It sharply reduces three failure modes at once: invalid tool calls (E1, $350\rightarrow116$, $-67\%$), redundant or unnecessary calls (E3, $448\rightarrow163$, $-64\%$), and, most tellingly, \emph{tool result not used}~(E4, $867\rightarrow532$, $-39\%$), which is precisely the behavior the tool reward targets, since it explicitly rewards conditioning subsequent reasoning on tool feedback. Error-free trajectories (E0) correspondingly grow from $644$ to $802$. In contrast, the higher-level semantic errors barely move: incorrect argument values (E2) go from $682$ to $717$, and incorrect reasoning from a valid tool output (E5) from $770$ to $811$, together becoming the two dominant residual failure modes after RL. In short, RL first fixes \emph{whether} tools are invoked well and \emph{whether} their outputs are consumed at all, whereas correctly grounding tool arguments and drawing the right inference from a valid observation remain the harder open problem, and the biggest lever for future work.

\end{document}